\documentclass[twoside,journey]{IEEEtran}
\usepackage{makecell}
\usepackage{hyperref}
\usepackage{array}
\usepackage{graphicx,amssymb,amsmath}
\usepackage{multicol}
\usepackage[noadjust]{cite}
\usepackage{setspace}
\usepackage{subfigure}
\usepackage{graphicx}
\usepackage{float}
\usepackage{url}
\usepackage{stfloats}
\usepackage{amsthm,pifont}
\usepackage{flushend}
\usepackage{cases,subeqnarray}
\usepackage{bm,multirow,bigstrut}
\usepackage{amsmath, amsthm, amssymb}
\usepackage{textcomp}
\usepackage{latexsym,bm}
\usepackage{booktabs}
\usepackage{xcolor}
\usepackage{mathtools}
\usepackage{dsfont}
\usepackage{extarrows}
\usepackage{epsfig}
\usepackage{epsfig}
\usepackage{epstopdf}
\usepackage[noend]{algpseudocode}
\usepackage{algorithmicx,algorithm}
\usepackage{xspace}
\usepackage{listings}
\usepackage{xcolor}
\definecolor{commentcolor}{RGB}{85,139,78}
\definecolor{stringcolor}{RGB}{206,145,108}
\definecolor{keywordcolor}{RGB}{0,0,128}
\definecolor{backcolor}{RGB}{220,220,220}
\theoremstyle{plain}

\theoremstyle{plain}
\newtheorem{rem}{Remark}

\newtheorem{prop}{Proposition}
\usepackage{amsmath}

\newtheorem{lemma}{Lemma}
\IEEEoverridecommandlockouts

\begin{document}

\title{Attention-aware Resource Allocation and QoE Analysis for Metaverse xURLLC Services}
\author{Hongyang~Du, Jiazhen~Liu, Dusit~Niyato,~\IEEEmembership{Fellow,~IEEE}, Jiawen~Kang, Zehui Xiong, Junshan~Zhang,~\IEEEmembership{Fellow,~IEEE}, and Dong~In~Kim,~\IEEEmembership{Fellow,~IEEE}
\thanks{H. Du and D. Niyato are with the School of Computer Science and Engineering, Nanyang Technological University, Singapore (e-mail: hongyang001@e.ntu.edu.sg, dniyato@ntu.edu.sg). J.~Liu is with the School of Information, Renmin University of China, China (e-mail: liujiazhen@ruc.edu.cn). J. Kang is with the School of Automation, Guangdong University of Technology, China. (e-mail: kavinkang@gdut.edu.cn). Z. Xiong is with the Pillar of Information Systems Technology and Design, Singapore University of Technology and Design, Singapore (e-mail: zehui\_xiong@sutd.edu.sg). J. Zhang is with the Department of Electrical and Computer Engineering, University of California Davis, USA (e-mail: jazh@ucdavis.edu). D. I. Kim is with the Department of Electrical and Computer Engineering, Sungkyunkwan University, South Korea (e-mail: dikim@skku.ac.kr).}

}
\maketitle
\vspace{-1cm}
\begin{abstract}
Metaverse encapsulates our expectations of the next-generation Internet, while bringing new key performance indicators (KPIs). Although conventional ultra-reliable and low-latency communications (URLLC) can satisfy objective KPIs, it is difficult to provide a personalized immersive experience that is a distinctive feature of the Metaverse. Since the quality of experience (QoE) can be regarded as a comprehensive KPI, the URLLC is evolved towards the next generation URLLC (xURLLC) with a personalized resource allocation scheme to achieve higher QoE. To deploy Metaverse xURLLC services, we study the interaction between the Metaverse service provider (MSP) and the network infrastructure provider (InP), and provide an optimal contract design framework. Specifically, the utility of the MSP, defined as a function of Metaverse users' QoE, is to be maximized, while ensuring the incentives of the InP. To model the QoE mathematically, we propose a novel metric named Meta-Immersion that incorporates both the objective KPIs and subjective feelings of Metaverse users. Furthermore, we develop an attention-aware rendering capacity allocation scheme to improve QoE in xURLLC. Using a user-object-attention level dataset, we validate that the xURLLC can achieve an average of $20.1\%$ QoE improvement compared to the conventional URLLC with a uniform resource allocation scheme. The code for this paper is available at \url{https://github.com/HongyangDu/AttentionQoE}.
\end{abstract}
\IEEEpeerreviewmaketitle
\section{Introduction}\label{Intro}	
\subsection{Background}
Initially, as a concept in fiction~\cite{joshua2017information}, Metaverse holds people's aspirations for the future world. The deployment of Metaverse services relies on the rapid advances of communication technologies. For example, Terahertz communications provide high data rate and low latency~\cite{du2022performance}, advanced network techniques improve the security of Internet of Things (IoT)~\cite{verma2021smart,verma2020network}, and multiple-input multiple-output (MIMO) technology further enhances the communication reliability. These promising techniques help to support key service areas in the fifth-generation (5G) network architecture such as massive machine-type communications, enhanced mobile broadband, and ultra-reliable low-latency communications (URLLC)~\cite{she2021tutorial,saad2019vision,she2020deep}, thereby providing the technical basis for the implementation of Metaverse services. In particular, URLLC has great potential to support a series of significant Metaverse services based on graphical techniques such as virtual reality (VR) and augmented reality (AR), e.g., virtual traveling and meeting~\cite{kelkkanen2020bitrate}. The reason is that these services require reliable transmission of user-object interactions and low latency delivery of virtual object data.

However, the emergence of Metaverse services and applications brings new key performance indicator (KPI) requirements that are different from those in the conventional URLLC services. Specifically, to provide users with an immersive experience in a deeply interactive environment of virtual and real worlds, graphical technology-enabled next-generation Internet services should have the ability to bring users personalized stimulation of various sense information. The ``user-centric'' requirement is difficult to be achieved solely by the conventional URLLC that focuses mainly on objective KPIs such as data rate and outage probability. This leads to the presentation of the next generation URLLC (xURLLC)~\cite{lu2022outlook} to fully consider the users' subjective feelings.

Therefore, novel yet sophisticated Metaverse xURLLC services need personalized service design, real-time interaction, and energy efficiency network availability, which places great demands on access technologies and novel user-centric resource allocation schemes~\cite{dong2020deep,han2021dynamic,she2021grand,du2022rethinking}. For example, Metaverse xURLLC services, e.g., AR game, should ensure high quality in environment sensing, data transmission, and personalized graphic rendering to provide users an excellent feeling of immersion. To consider comprehensively the above multiple technical indicates and user feelings, we model the QoE to encompass significant KPIs in Metaverse xURLLC services, and take QoE as the service design goal.

To maximize QoE, we need to personalize the design of the resource allocation scheme according to the users' interest difference. This idea is also inspired by the recent successes in semantic communications~\cite{yang2022semantic}. By extracting and processing the semantic information from the data to be transmitted, several service KPIs can be greatly improved under the same resource constraints~\cite{du2022semantic}. The reason is that more transmission resources can be allocated to the data related to the tasks. In Metaverse xURLLC services such as virtual traveling, meeting, and sports, the task of the Metaverse service provider (MSP) could be {\textit{providing users excellent Metaverse immersion experiences}}. Thus, by analyzing and predicting users' interests, we can allocate more network resources, e.g., rendering capacity, to virtual objects in which users will pay more attention, just as we allocate more transmission resources to the task-related data in semantic communications. Here we use the term ``user-object-attention value'' to express the user's interest in the virtual object.
To implement the above Metaverse xURLLC service scheme, the following questions needs to be answered:
\begin{enumerate}
\item[{\textbf{Q1)}}] How to implement the designed Metaverse xURLLC service scheme in the practical service market?
\item[{\textbf{Q2)}}] How to quantify QoE using representative KPIs and reflect the differences in user-object-attention values?
\item[{\textbf{Q3)}}] How can the MSP predict user attention values to virtual objects for resource allocation algorithm design?
\end{enumerate}
\begin{figure*}[t]
\centering
\includegraphics[width=0.85\textwidth]{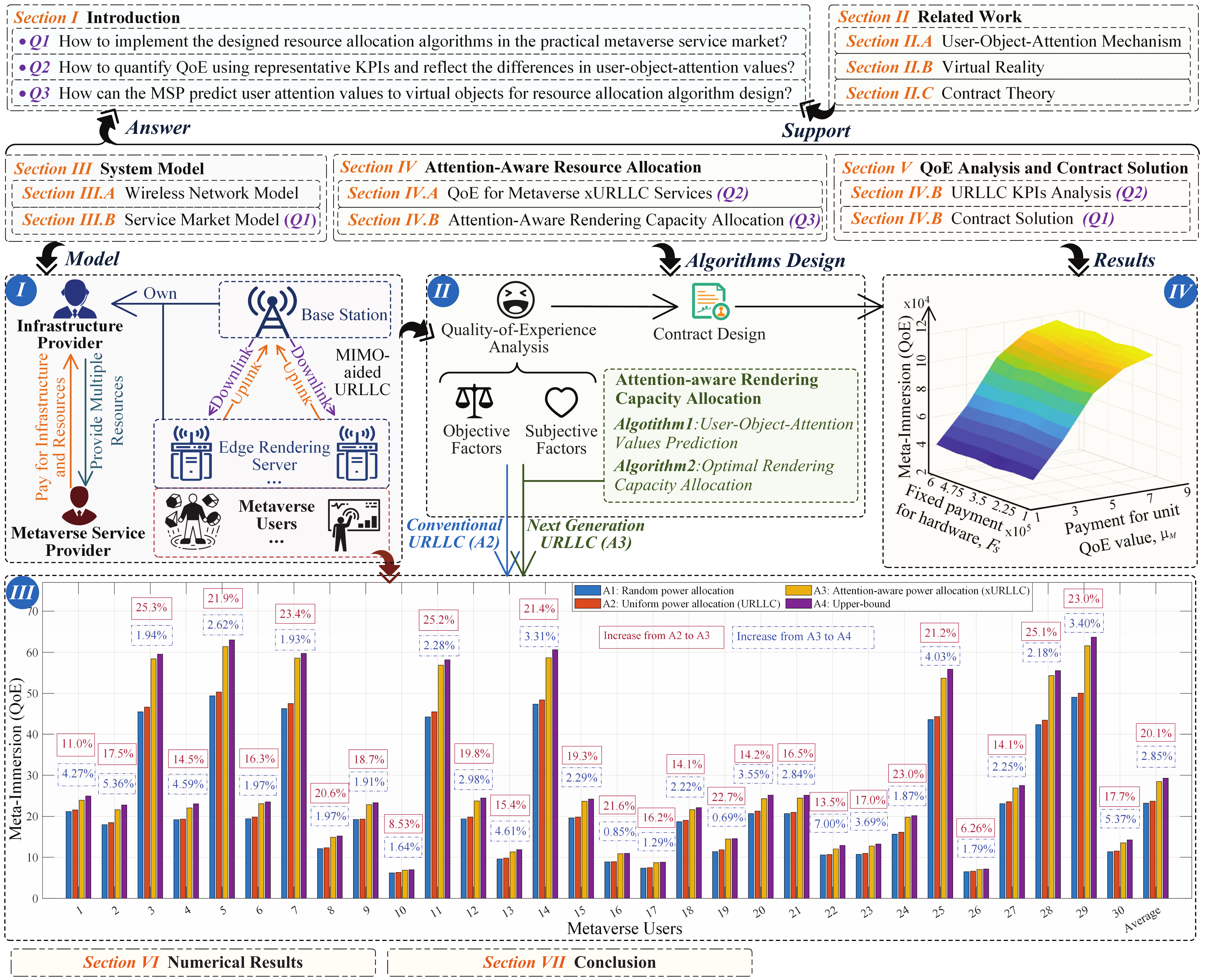}
\caption{Structure and main contributions of this paper. Part I shows the system model. Part II shows the ideas of QoE formulation and algorithms design. Part III shows that the Meta-Immersion of $30$ users under three different resource allocation schemes, i.e., random, uniform (in conventional URLLC without considering the attention differences among users), attention-aware rendering capacity allocation (in xURLLC), and the upper-bound obtained from the ground truth. Part IV shows the Meta-Immersion versus the payment from the MSP to the network infrastructure provider (InP) for unit QoE value, $u_M$, and the fixed payment from the MSP for using the hardware infrastructures, $F_s$.}
\label{NewModel}
\end{figure*}
To fill the aforementioned research gaps, we study the service market among the MSP, Metaverse users, and the InP, as shown in Fig.~\ref{NewModel} (Part I). Because the Metaverse services are emerging businesses, the InP gains new opportunities to sell network resources to the MSP. From the MSP's perspective, it ``employs'' the InP to provide virtual services to users. Therefore, a qualified mechanism should be designed to reward the InP comprehensively for driving the InP's incentives to take actions. Specifically, to maximize the utility of the MSP and ensure the InP agrees with the incentive scheme, we have to consider the incentive compatible (IC) constraint~\cite{zhang2017contract}, which means that the InP provides the optimal amount of resources to maximize its own utility, and the individual rationality (IR) constraint~\cite{zhang2017contract}, which means that the utility of the InP is larger than a threshold. To this end, contract theory can be a suitable solution for the incentive scheme design problem~\cite{zhang2017multi}.

\subsection{{\color{black} Contributions}}
We aim to design an MSP-InP contract\footnote{We focus on the contract design between one MSP and one InP~\cite{zhang2017multi,zhang2017contract}. For the multiple InPs case, because the MSP can assign different services to different InPs, it is still an one-to-one contract design problem~\cite{zhang2017multi}. The case that multiple InPs compete one Metaverse service is left to future work.} that weighs the InP's contributions by the QoE and maximizes the utility of the MSP. 
The main contributions of this paper are summarized as follows:

{\color{black} \begin{itemize}
		\item We introduce a new xURLCC service framework for Metaverse. By analyzing the physical edge network and the service market between the MSP and the InP, we propose a contract theory-based framework that takes both multidimensional network resources and the QoE into account. The utility of the MSP is maximized under the service market constraints while ensuring that the InP obtains a satisfactory incentive to participate in the contract (for ${\textbf{Q1}}$).
		\item For the proposed framework, we then design a novel metric named Meta-Immersion (MI) to model the QoE from the perspective of Metaverse users. By using the psychological Weber–Fechner Law, MI includes both objective service quality and subjective feelings of users. The MIMO technique is used to o achieve URLLC. We then derive the closed-form expressions of downlink data rate and uplink bit error probability (BEP) to obtain MI expression in the xURLLC system (for ${\textbf{Q2}}$).
		\item With the MI metric, we further propose an attention-aware rendering capacity allocation algorithm to achieve xURLLC, which can predict users' attention to all virtual objects through the historical sparse user-object-attention records and then allocate resources optimally (for ${\textbf{Q3}}$).
		\item To facilitate the quantification of QoE in a Metaverse context, we analyze the user-object-attention level (UOAL) dataset~\cite{du2022exploring} that contains the attention values of $30$ users to $96$ objects. Using UOAL, we validate that the proposed xURLLC attention-aware allocation scheme can improve the QoE averagely by $20.1\%$ compared to the conventional URLLC with the uniform resource allocation scheme. A higher percentage of QoE improvement, e.g., $40\%$, is achieved when the total resources for Metaverse xURLLC services are limited.
\end{itemize}}
The structure of this paper is shown in Fig.~\ref{NewModel}. Section~\ref{rela} reviews the related work in the literature. In Section~\ref{S2}, we introduce the system model, which contains the wireless MIMO network and the service models. In Section~\ref{S5}, the QoE metric, i.e., MI, is formulated and the attention-aware resource allocation algorithms are proposed to achieve the xURLLC. In Section~\ref{SR5}, we derive the closed-form MI and obtain the contract solution. Section~\ref{nums} shows the effectiveness of our proposed schemes. Section~\ref{S7} concludes this paper.
\section{Related Work}\label{rela}
This section provides a brief review of the related work on the user-object-attention mechanism, VR, and contract theory in the service market.

\subsection{User-Object-Attention Mechanism}
Eye-movements, the reliable mirror of attention allocation, are second nature of humans \cite{schutz2011eye}. The authors in \cite{onat2014contributions} find that, in the free-viewing of natural scenes, fixation duration is dependent mainly on attention ratings.
Thanks to recent development in the field of head-mounted displays (HMDs) and computer graphics, pervasive eye trackers can be used easily to collect eye movement data \cite{bozkir2019assessment,berkovsky2019detecting,braunagel2017online}. A large amount of eye movement data is indeed useful for the personalized service design \cite{bozkir2019assessment}. In general, it has been shown that the eye movement data can be employed for the assessment of situational attention \cite{bozkir2019assessment}, detection of personality traits \cite{berkovsky2019detecting}, and activity recognition even in challenging daily tasks \cite{braunagel2017online}. Thus, the user-object-attention values can be obtained from eye movement records and help us make better use of rendering resources to achieve the xURLLC.

\subsection{Virtual Reality}\label{resolution}
As one of the most significant enablers of 6G communications, VR demonstrates great potential as a key technology to access the virtual world, e.g., Metaverse~\cite{saad2019vision}. With the rise of consumer-level VR devices, and especially HMDs with lower graphical compute capabilities, one obstacle, which is hard to be solved in conventional URLLC, is to achieve high virtual object quality at a low cost. To solve this problem, the authors in \cite{feng2013low} optimize video streaming with the help of eye-tracking. A hidden Markov model is used to predict the user's gaze region that is then encoded in higher quality than the rest. It is shown that 29\% bitrate savings can be achieved without the users reporting quality degradation. Similarly, the authors in \cite{ghinea2009eye} adopt the virtual content in real-time according to the user's gaze. However, allocating resources based on user gaze behavior requires real-time computing and ignores the interest of users, i.e., attention to virtual objects. Thus, a better option is to allocate rendering capacity, e.g., in the form of resolution, based on the user's attention to different virtual objects. In fact, some literature discusses the adaptive resolution method in VR. A mechanism that can adjust the VR resolution according to task complexity is proposed in~\cite{su2019development}. Moreover, the authors in \cite{likamwa2021adaptive} discuss the adaptive resolution-based trade-offs for energy-efficient visual computing systems.

\subsection{Contract Theory}
Contract theory has been widely used in wireless networks~\cite{zhang2017survey}. To the best of our knowledge, the authors in \cite{gao2011spectrum} first apply contract theory to the spectrum sharing problem. Since then, the contract theory is used in a growing number of problems, e.g., D2D communications~\cite{zhang2015contract}, data transactions~\cite{zhang2015contract}, and energy management~\cite{lv2021contract}. The book \cite{zhang2017contract} studies and summarizes the use of contract theory in wireless networks, and discusses the multi-contract design and the one-to-one contract optimization between two entities. The latter is used as the incentive mechanism in crowd-sourcing to maximize the utility of the principal while providing the users continuous incentives \cite{zhang2017multi}. However, the cooperation patterns are impacted by the recent development of wireless networks, in which multiple resources can be used~\cite{wang2022joint}. Furthermore, the optimal contract design is decided by the QoE. The contract design between the MSP and the InP of the wireless network has not been fully investigated.

Inspired by these existing works, we apply contract theory to study the cooperation between the MSP and the InP in the Metaverse service market, and design an attention-aware resource allocation scheme to maximize the QoE.

\section{System Model}\label{S2}
With the help of wireless network infrastructures owned by the InP, the MSP can rapidly deploy Metaverse applications and services to improve the QoE of users. We consider one of the basic Metaverse xURLLC services: {\it Providing users with virtual immersion experiences}, such as virtual traveling and meeting~\cite{um2022travel}. In this kind of services, there are several options for users, e.g., different virtual traveling scenarios. In this section, we first present the wireless network for Metaverse xURLLC services, and then discuss the service market model between the MSP and the InP.
\subsection{Wireless Network Model}\label{MTN}
In wireless MIMO networks, multiple antennas are adopted at both the transmitter and receivers to obtain considerable array gains and improve the channel quality. As shown in Fig. \ref{NewModel} (Part I), we consider that a multi-antenna cloudlet-enabled base station (CBSs) can use the wireless communication resources, i.e., transmit power and bandwidth, to interact with the edge devices. Multi-antenna edge devices, such as rendering servers (RSs), receive the downlink virtual object data and render objects for users.

{\color{black}Wireless environments have a significant impact on Metaverse xURLLC service quality. Strong small-scale fading and severe interference from other co-channel users result in the InP using more resources to maintain the same QoE as when the wireless environment is favorable. Here we study the wireless connections between CBSs and RSs, considering the effects of number of antennas, small-scale and large-scale fading, transmit power, number of co-channel interference paths, and interference power.} Because the RS can be placed in the user's home or integrated into the HMD, we consider one RS serves one user \cite{tsao2021technical}. With the network parameters shown in Table~\ref{Networkparameters}, we express the probability density function (PDF) of the signal-to-interference ratio (SIR) under the interference-limited scenario for the $k_{\rm th}$ $\left(k = 1,\ldots,M_U\right) $ user as~\cite{zhang2012statistical}
\begin{equation}\label{PDF}
{f_{{\gamma _{k}}}}\!\left( \gamma  \right) \!=\! \frac{{{\Lambda _{k}}{{\left( {\gamma {\Lambda _{k}}} \right)}^{{M_C}{M_U} - 1}}}}{{B\!\left( {{M_C}{M_U},{M_C}{N_Q}} \right)\!{{\left( {1\! + \!\gamma {\Lambda _{k}}} \right)}^{{M_C}{M_U} + {M_C}{N_Q}}}}},
\end{equation}
where
\begin{equation}\label{delk}
{\Lambda _{k}} \triangleq \frac{{{P_{k,q}}{\mu _{k,q}}}}{{{M_C}\zeta_{k} {D_{k}^{ - {\alpha _{k}}}} {P_{k}^{\left(D\right) }}{\mu _{k}}}},
\end{equation}
\begin{equation}\label{agfeaga}
\zeta_{k} = \mathbb{E}\left[{\lambda_{\rm max}}\right]/\mathbb{E}\left[{{\sum\limits_{i = 1}^{\min \left\{ {{M_C},{M_U}} \right\}} {{\lambda _i}} }}\right],
\end{equation}
$ {{\lambda _i}} $ {\small $\left( i = 1, \ldots ,\min \left\{ {{M_C},{M_U}} \right\}\right) $} are the non-zero eigenvalues of the matrix {\small $ {\bm H}_{k}^H {{\bm H}_{k}} $}, $\lambda_{\rm max}$ is the largest eigenvalue, {\small $\mathbb{E}\left[{\cdot}\right]$} represents statistical expectation, and ${B \! \left( { \cdot , \cdot } \right)}$ is the Beta function~\cite[eq. (8.384.1)]{gradshteyn2007}. With the help of \eqref{PDF}, we mathematically analyze the objective service KPIs in Section \ref{SectionQoE} which impact the QoE of Metaverse users and the utilities of the InP and the MSP.

The wireless network architecture leads to a new market value chain comprising the InP, the MSP, and users. The InP's input of physical infrastructures, computing, and communication resources helps the MSP deploy Metaverse xURLLC services to users, and the MSP needs to pay the InP.

\begin{table}
\caption{Wireless Communications Network Parameters.}
\label{Networkparameters}
\centering
\renewcommand{\arraystretch}{1}
{\small\begin{tabular}{m{3cm}<{\centering}|m{4cm}<{\centering}}
\hline
\textbf{Notation} & \textbf{Network Parameter} \\
\hline
$M_C$ &  Number of antennas in CBS \\
\hline
$M_{U}$ & Number of antennas in RS  \\
\hline
$N_U$ & Number of Metaverse users  \\
\hline
$N_Q$ & Number of co-channel interference paths \\
\hline
$D_{k}$ & Transmission distance between the $j_{\rm th}$ CBS and $k_{\rm th}$ user  \\
\hline
$\alpha_{k}$ & Path loss exponent between the $j_{\rm th}$ CBS and $k_{\rm th}$ user  \\
\hline
$P_{k}^{\left(D\right) }$ & Downlink transmit power  \\
\hline
$P_{k,q}$ & Interference power  \\
\hline
${\bm H}_{k} \in {\mathbb{C}^{M_{U} \times M_{C}}}$ & Channel between the $k_{\rm th}$ user and the $j_{\rm th}$ CBS  \\
\hline
$\mu_{k}$ & Rayleigh channel coefficient of data links  \\
\hline
$\mu_{k,q}$ & Rayleigh channel coefficient of interference links  \\
\hline
\end{tabular}}
\end{table}

\subsection{Service Market Model}\label{S3}
We study the contract theory-based payment plan in this section. The utilities of the InP and the MSP are formulated respectively, by considering the QoE of Metaverse users.
\begin{figure}[t]
\centering
\includegraphics[width=0.4\textwidth]{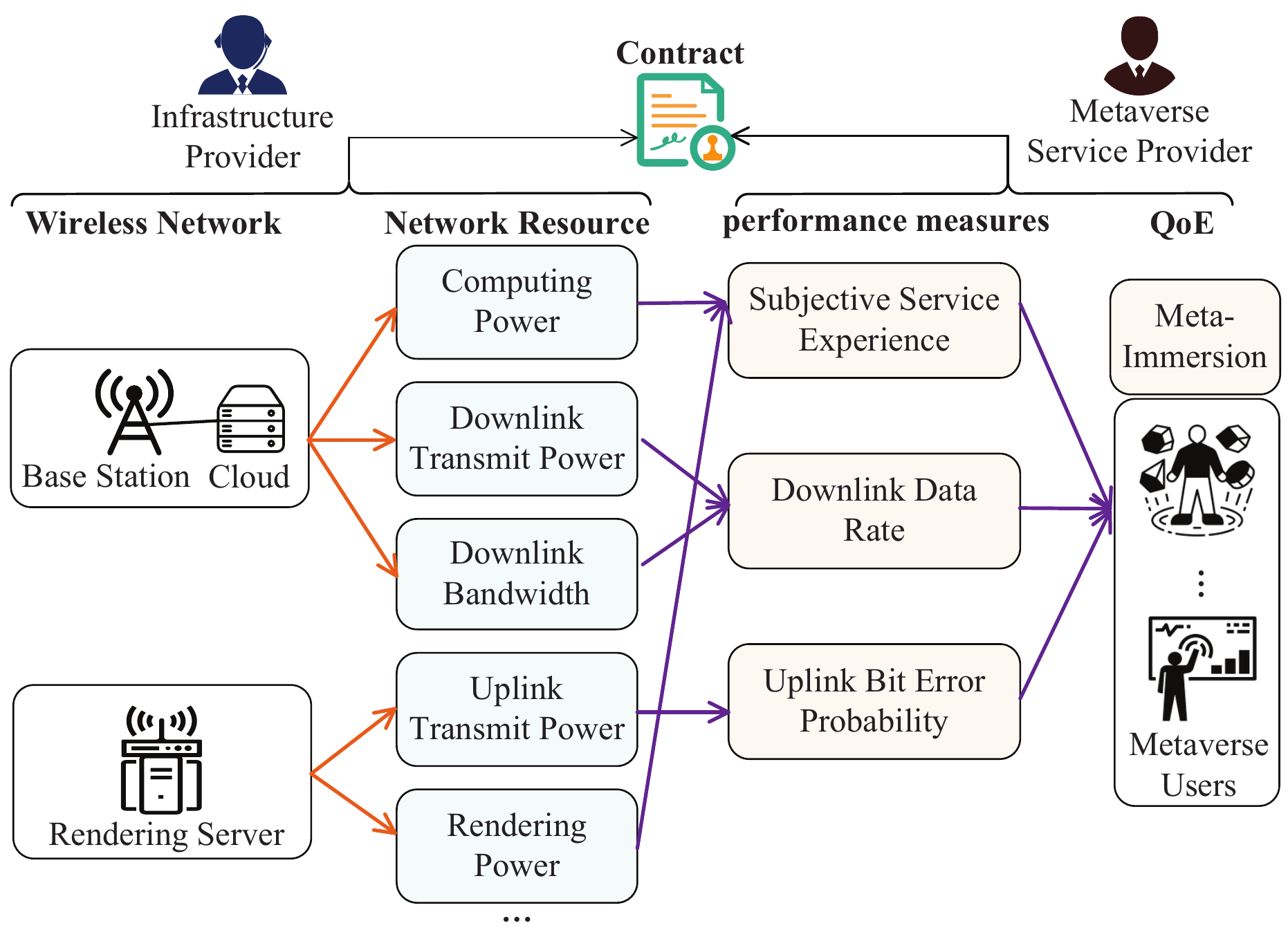}%
\caption{A contract theory-based Metaverse xURLLC service market with multi-dimension resources and the QoE-based payment.}
\label{ConT}
\end{figure}
\subsubsection{Payment Plan}
To realize the full potential of the wireless network, an appropriate payment plan is required, which allows all stakeholders and Metaverse xURLLC service users to benefit from the cooperation.
We propose a contract theory-based payment plan, in which the InP receives the fee from the MSP according to the computing and communication resources that the InP provides and the QoE of MSP's users. Thus, the revenue function of the InP can be expressed as
\begin{equation}\label{utilityInP}
{I_{{\rm{InP}}}} = {F_s} + {u_M}\sum\limits_{i = 1}^{N_U} {{{\mathcal M}_{i}}},
\end{equation}
where {\small ${F_s}$} is the fixed payment from the MSP for using the hardware infrastructures, $u_M$ is the fee for unit QoE value, and {\small ${{\mathcal M}_{i}}$} is the value of the quantified QoE of $i_{\rm th}$ Metaverse user. Note that {\small ${{\mathcal M}_{i}}$} is determined by various resources of the InP as well as by the user's personalized subjective parameters, which is discussed in detail in Section \ref{SectionQoE}.

\subsubsection{Utility of the InP}\label{UINPS}
As shown in Fig. \ref{ConT}, the InP invests multidimensional network resources, i.e., downlink transmit power {\small $P_k^{\left(D\right) }$}, downlink bandwidth {\small $B_k$}, uplink transmit power {\small $P_k^{\left(U\right) }$}, and rendering capacity {\small $P_{k}^{\left( R\right) }$}, which are denoted by
\begin{equation}
{\bm \Theta}  = \left( {P_k^{\left(D\right) }, B_k , P_k^{\left(U\right) }, P_{k}^{\left( R\right) }} \right).
\end{equation}
Then, the InP's cost is defined in a quadratic form as {\small $ {\left\| {{{\bm u}_\Theta } \odot \Theta } \right\|^2} $}, where $\odot$ means the Hadamard product~\cite{horn1990hadamard}, ${{\bm u}_\Theta } = \left( {\sqrt{u_{1\theta }}, \ldots, \sqrt{u_{{N_\theta }\theta }}} \right)$, and $u_{{i}\theta }$ is the cost for the $i_{\rm th}$ unit resource \cite{zhang2017contract}. 
These resources impact the KPIs of the Metaverse xURLLC services. Here we study three types of KPIs, i.e., downlink data rate {\small $R^{\left(D\right)}_{k}$}, uplink BEP {\small $E_k^{\left(U\right)} $}, and subjective service experience {\small $X_k$}, which are denoted by
\begin{equation}\label{KPI}
\Psi  = \left( {R^{\left(D\right)}_{k}, E_k^{\left(U\right)} ,X_k} \right).
\end{equation}
{\color{black}Note that {\small $X_k$} indicates the subjective feeling of users towards xURLLC service that is discuss in Section \ref{SectionQoE}. {\small $R^{\left(D\right)}_{k}$} and {\small $E_k^{\left(U\right)}$} are objective conventional URLLC KPIs that are considered due to the following two facts:
\begin{itemize}
	\item The VR data delivery requires high downlink data rate to ensure low transmission latency.
	\item For the data uplink, the data amount is not large, but small bit error probability should be ensured to achieve accurate user-object interactions~\cite{yang2021feeling}.
\end{itemize}
Our analysis framework can be extended easily when other types of KPIs are considered.
}

The Metaverse xURLLC service is an emerging business with uncertainty, which may lead to the situation that the InP is not certain to be profitable from cooperation with the MSP. Specifically, the demands of users depend highly on socio-economic factors such as the popularity of the MSP among their friends \cite{nazir2008unveiling}. This motivates us to consider the user risk preference~\cite{tremewan2020behavioral} in the InP utility. A measure of risk aversion that is commonly used in financial economics is the Arrow–Pratt measure of relative risk aversion (RRA)~\cite{tremewan2020behavioral}, as
\begin{equation}\label{rra}
{\rm RRA} =  - \frac{{U_{{\rm{InP}}}^{''}\left( {{W_{{\rm{InP}}}}} \right)}}{{U_{{\rm{InP}}}^{'}\left( {{W_{{\rm{InP}}}}} \right)}}{W_{{\rm{InP}}}},
\end{equation}
where 
\begin{equation}\label{winp}
W_{{\rm{InP}}} \triangleq I_{\rm InP}-{\left\| {{{\bm u}_\Theta } \odot \Theta } \right\|^2},
\end{equation}
is the difference between the revenue and cost, and {\small $ {U_{{\rm{InP}}}}\left( {{W_{{\rm{InP}}}}} \right) $} is the utility of the InP. The RRA measures the degree of risk aversion of the InP under consideration, where a larger RRA means more risk aversion. Here we consider that the InP has constant relative risk aversion (CRRA) preferences for the Metaverse services. Then, a common utility function in economics with CRRA is the power utility as~\cite{tremewan2020behavioral}
\begin{equation}\label{utilitINP}
{U_{{\rm{InP}}}}\left( {{W_{{\rm{InP}}}}} \right) = \frac{{{W_{{\rm{InP}}}}^{1 - \tau }}}{{1 - \tau }},
\end{equation}
where $\tau$ is the value of RRA when substituting \eqref{utilitINP} into \eqref{rra}. Typically, the contract between the MSP and the InP is designed to ensure that the utility of the InP is larger than a certain threshold. Thus, we consider $0 \le \tau < 1$. Note that, for $\tau = 0$, we can rewrite \eqref{utilitINP} as {\small $ {U_{{\rm{InP}}}}\left( {{W_{{\rm{InP}}}}} \right) = {W_{{\rm{InP}}}} $}, which means that the preference of the InP shows risk neutrality.

\subsubsection{Utility of MSP}
Since the MSP is typically a well-established business, e.g., Facebook, and is more tolerant to the uncertainties, we consider that the MSP is risk neutral, and express the utility of the MSP as
\begin{equation}\label{umsp}
{U_{\rm MSP}} = \sum\limits_{i = 1}^{{N_U}} {\left( {\omega_{Ui}}+{\mu_{Ui}}{{\mathcal M}_{i}}\right) }- {I_{{\rm{InP}}}},
\end{equation}
where ${\omega _{Ui}}$ represents the basic fee paid by the $i_{\rm th}$ user for access to Metaverse, and ${\mu_{Ui}}$ denotes the additional fee of virtual services, e.g., the user pays for the high-quality access service to obtain higher QoE~\cite{du2021optimal}.
\begin{rem}
We observe that the utilities of both InP and MSP increase if the QoE of Metaverse users can be increased under the same resource investment. The reason is that the InP can gain more payments at the same amount of resources cost, and the MSP can attract more users by providing higher QoE.
\end{rem}
As such, we now analyze the QoE from the rendering quality perspective, i.e., {\small $X_k$} in \eqref{KPI} denotes the rendering quality that users perceive. In Metaverse, through VR access, a user will perceive and put attention to different displayed objects differently. For example, the user may have more attention to persons' avatars rather than a table in a virtual meeting. Thus, it is more efficient to use higher rendering capacity on objects of greater interest to the user. To facilitate such resource allocation, we predict the user's attention to objects to be displayed in Section~\ref{S5}, using the sparse historical records of objects seen by the user before.

\subsubsection{Optimal Contract Design}\label{sconter}
Combing \eqref{utilitINP}, \eqref{umsp}, \eqref{utilityInP}, and \eqref{winp}, we can express the utility functions of the MSP and the InP as
\begin{equation}\label{reutilityMSP}
{U_{\rm MSP}}\left( {{\bm{\Theta }},{F_s},{u_M}} \right) \!=\! \sum\limits_{i = 1}^{{N_U}} {\left( {{\omega _{Ui}} \!+ \!\left( \mu _{Ui}\!-\!{u_M}\right) {{\cal M}_i}} \right)} \! -\! {F_s},
\end{equation}
and
\begin{equation}\label{reutilityInP}
{U_{{\rm{InP}}}}\!\left(\! {{\bm{\Theta }},\!{F_s},\!{u_M}} \!\right) \!=\! \frac{1}{{1 \!- \!\tau }}\!{\left(\!\! {{F_s}\!\! + {u_M}\!\!\sum\limits_{i = 1}^{{N_U}} {{{\cal M}_i}} \! - {{\left\| {{{\bm{u}}_\Theta }\Theta } \right\|}^2}} \!\!\right)^{1 - \tau }},
\end{equation}
respectively. The contract provided by the MSP includes two items, i.e., {\small $ \left\{ {{F_s},{u_M}} \right\} $}. To design the optimal contract, we formulate the MSP's utility maximization problem while providing the InP with the necessary incentives to agree on the contract. The optimization problem can be expressed as~\cite{zhang2017multi}
\begin{equation}
\begin{array}{*{20}{c}}
{\mathop {\max }\limits_{{\bm{\Theta }},{F_s},{u_M}} }&{{U_{\rm MSP}} \left( {{\bm{\Theta }},{F_s},{u_M}} \right)}\\
{\rm{s.t.}}&{\left\{ \begin{array}{l}
{{\bm{\Theta }}^*} \in \arg \mathop {\max }\limits_{\bm{\Theta }} {U_{{\rm{InP}}}}\left( {{\bm{\Theta }},{F_s},{u_M}} \right),\\
{U_{{\rm{InP}}}}\left( {{{\bm{\Theta }}^*},{F_s},{u_M}} \right) \ge U_{\rm th}^{\rm InP},
\end{array} \right.}
\end{array}
\end{equation}
where the first constraint is the IC constraint~\cite{zhang2017contract}, i.e., {\small ${P_k^{\left(D\right) }}^*$}, {\small ${B_k}^*$}, {\small ${P_k^{\left(U\right) }}^*$}, and {\small ${P_{k}^{\left( R\right) }}^*$} is set to maximize its own utility. The second is the IR constraint~\cite{zhang2017contract} with a utility threshold {\small $U_{\rm th}^{\rm InP}$}.

This contract design problem can be viewed as a leader-follower game model, which can be easily solved if we know the value of ${\bm{\Theta }}$ under each {\small $ \left\{ {{F_s},{u_M}} \right\} $}~\cite{pang2005quasi}. Here, ${\bm{\Theta }}$ is designed to maximize the utility of the InP, which is equivalent to maximizing ${{{\cal M}_i}}$. Thus, we need to obtain the closed-form expression for the QoE of the $i_{\rm th}$ user ({\textbf{Q2}} in Section~\ref{Intro}), i.e., ${{{\cal M}_i}}$, and an optimal resource allocation scheme that maximizes ${{{\cal M}_i}}$ to provide high-quality Metaverse xURLLC services ({\textbf{Q3}} in Section~\ref{Intro}).

\section{Attention-Aware Resource Allocation}\label{S5}
In this section,  we propose a novel metric termed MI using the Weber-Fechner Law~\cite{dehaene2003neural}, to model the user's QoE in Metaverse xURLLC service. To help the InP make effective use of resources, we propose a two-step attention-aware rendering capacity allocation algorithm. 

\subsection{QoE Analysis for Metaverse xURLLC Services}\label{SectionQoE}
\begin{figure}[t]
\centering
\includegraphics[width=0.4\textwidth]{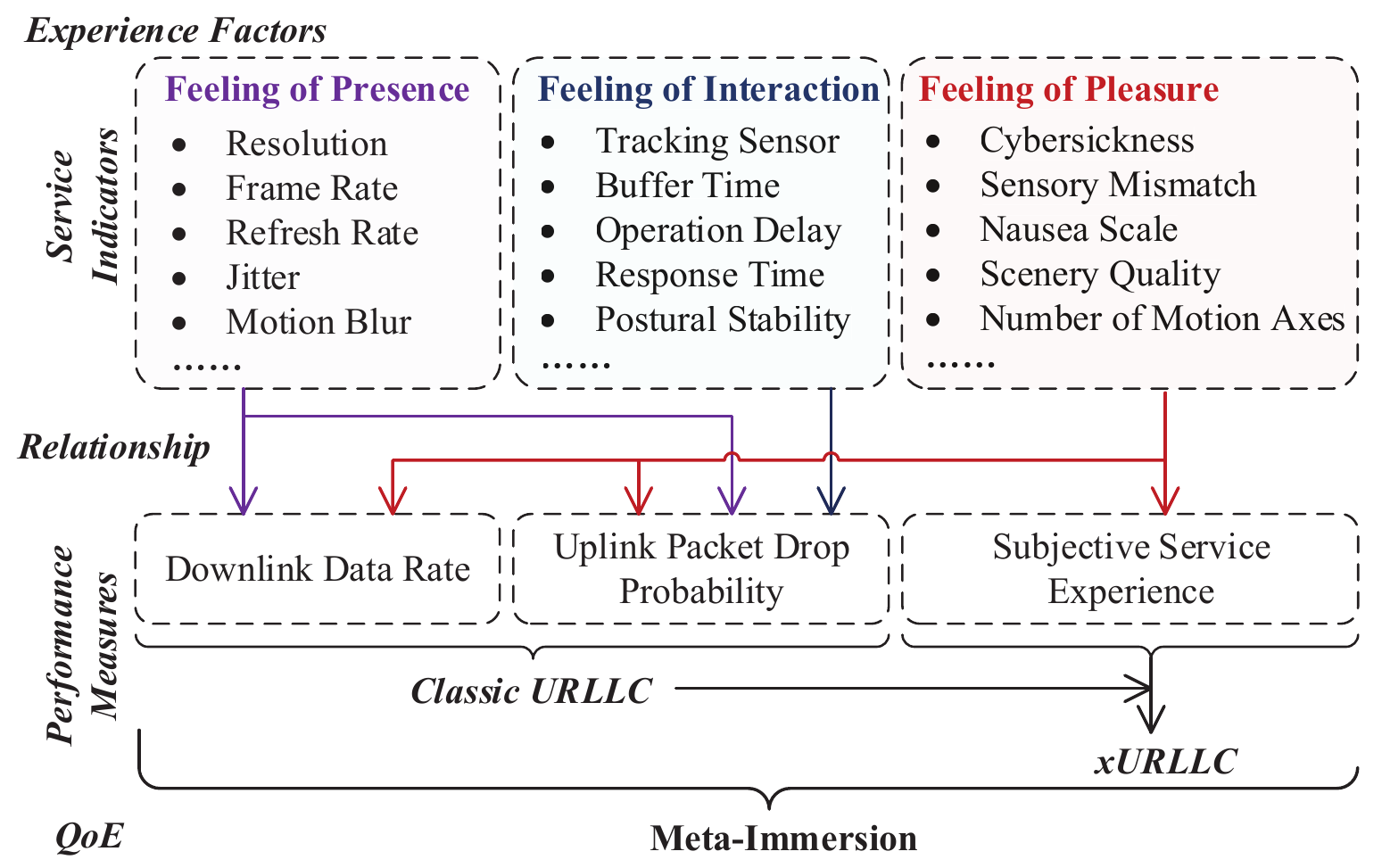}%
\caption{A novel performance metric in Metaverse: Meta-Immersion, and corresponding experience factors, service indicators, and KPIs, a.k.a., performance measures.}
\label{NemI}
\end{figure}
In the Metaverse service, three feelings, i.e., the feeling of presence, feeling of interaction, and feeling of pleasure, are important to the users' engagement~\cite{bouchard2008anxiety}. As shown in Fig. \ref{NemI}, to quantify these feelings and corresponding service indicators, we consider three KPIs, i.e., downlink data rate {\small $R^{\left(D\right)}_{k}$}, uplink BEP {\small $E_k^{\left(U\right)} $}, and subjective service experience, to present the formulation of the QoE.

Note that downlink data rate and uplink BEP are objective KPIs in conventional URLLC that impact the interaction between the physical and virtual worlds, and the subjective service experience indicates the perceived quality of Metaverse service. For Metaverse xURLLC service studied in this paper, e.g., virtual traveling, we consider that the subjective service experience is decided by the rendering quality. Thus, let {\small ${P_{n,k}^{\left( R \right)}}$} denote the rendering capacity and {\small $N_{Ok}$} denote the number of virtual objects that the $k_{\rm th}$ user sees in one service. The subjective service experience is a function of {\small ${P_{n,k}^{\left( R \right)}}$} and {\small $N_{Ok}$}.

{\color{black} To establish the relationship between the network performance and the subjective experience, we use the Weber–Fechner Law to derive the QoE, i.e., MI, as the {\textit{connection coefficient}} multiplied by the {\textit{logarithm of stimulus intensity}}~\cite{reichl2013logarithmic,reichl2010logarithmic,lubashevsky2019psychophysical}. In Metaverse virtual services, the stimulus intensity can be expressed in terms of the rendering capacity. The {\textit{connection coefficient}} is decided by the URLLC KPIs. Then, the differential of the MI of the $k_{\rm th}$ user can be expressed as
	\begin{equation}\label{WFL}
		{\rm d}{\cal M}_k = C_k \sum\limits_{n = 1}^{{N_{Ok}}} {{K_{n,k}}\frac{{{\rm{d}}{{P_{n,k}^{\left( R \right)}}}}}{{{{P_{n,k}^{\left( R \right)}}}}}} ,
	\end{equation}
	where {\small ${K_{n,k}}$} is the user-object-attention value that is a constant determined by users' physiological mechanism and subjective attention, $C_k$ is the {\textit{connection coefficient}}. As we discussed in Section~\ref{UINPS}, we consider the downlink data rate and the uplink BEP as the conventional URLLC KPIs. Thus, $C_k$ can be expressed as
 \begin{equation}
 	C_k = {\cal T}{\left(R^{\left(D\right)}_{k}\right)} {\cal T}{\left(1-E_k^{\left(U\right)}\right)}.
 \end{equation}
	The function {\small $ {\cal T}\left(  \cdot  \right) $} is used to eliminate the effect of the magnitudes~\cite{nishio2013service}, which is defined as 
	\begin{equation}\label{calT}
		{\cal T}\left( t \right) = \frac{{t - {t_{\min }}}}{{{t_{\max }} - {t_{\min }}}},
	\end{equation}
	where $ {{t_{\min }}} $ is the minimal threshold, $ {{t_{\max }}} $ is the maximal value that InP can provide. From \eqref{WFL}, we can observe that, unlike traditional psychological models, the stimulus perceived by the user is from Metaverse, and the transformation from the stimulus to the user's subjective feelings is influenced by the interaction between the virtual and the real worlds. 

By solving \eqref{WFL}, we obtain the MI for the $k_{\rm th}$ user as
\begin{equation}\label{mi}
{{\cal M}_k} = {\cal T}\!\left( {R_k^{\left( D \right)}} \right){\cal T}\!\left( {1 - E_k^{\left( U \right)}} \right)\sum\limits_{n = 1}^{{N_{Ok}}} {{K_{n,k}}\ln \left( {\frac{{P_{n,k}^{\left( R \right)}}}{{P_{\rm th}^{\left( R \right)}}}} \right)} ,
\end{equation}
where {\small $ {P_{\rm th}^{\left( R \right)}} $} is the minimal rendering capacity threshold, e.g., the resolution that is set for the object with the lowest attention. 

	\begin{rem}
	Different Metaverse URLLC services have different focus on KPIs. For example, in instant virtual meeting services, latency could be a significant performance metric. Fortunately, our proposed QoE metrics can be adapted to different service design requirements by adjusting for the {\textit{connection coefficient}}, i.e., $C_k$. For example, when latency, i.e., $L_k$, is considered in the QoE modelling, we can multiply ${\cal T}\left(L_{\textrm{max}} - L_k \right)$ in $C_k$, where $L_{\textrm{max}}$ is the maximum delay that Metaverse service can tolerate.
	\end{rem}
	\begin{rem}
		In conventional URLLC, the goal of network service design is to achieve optimal objective performance metrics, such as higher downlink data rate and lower BEP. Although ${\cal T}\!\left( {R_k^{\left( D \right)}} \right){\cal T}\!\left( {1 - E_k^{\left( U \right)}} \right)$ in \eqref{mi} can approach $1$, the rendering capacity is typically allocated uniformly without utilizing the user's attention mechanism in Metaverse services, which limits the further improvement of QoE. Thus, in xURLLC, unlike the uniform allocation scheme in URLLC, we design the personalized attention-aware rendering capacity allocation scheme as follows.
\end{rem}}


\subsection{Attention-Aware Rendering Capacity Allocation}\label{SectionSemtic}
There are two steps in the optimal rendering capacity allocation scheme. We first predict the user-object-attention values, i.e., {\small ${K_{n,k}}$}. Next, we design the optimal rendering capacity allocation scheme to maximize the MI given with predicted {\small ${K_{n,k}}$}. In the following, we design algorithms for the two steps, respectively.

\subsubsection{User-Object-Attention Values Prediction}
\begin{figure}[t]
\centering
\includegraphics[width=0.35\textwidth]{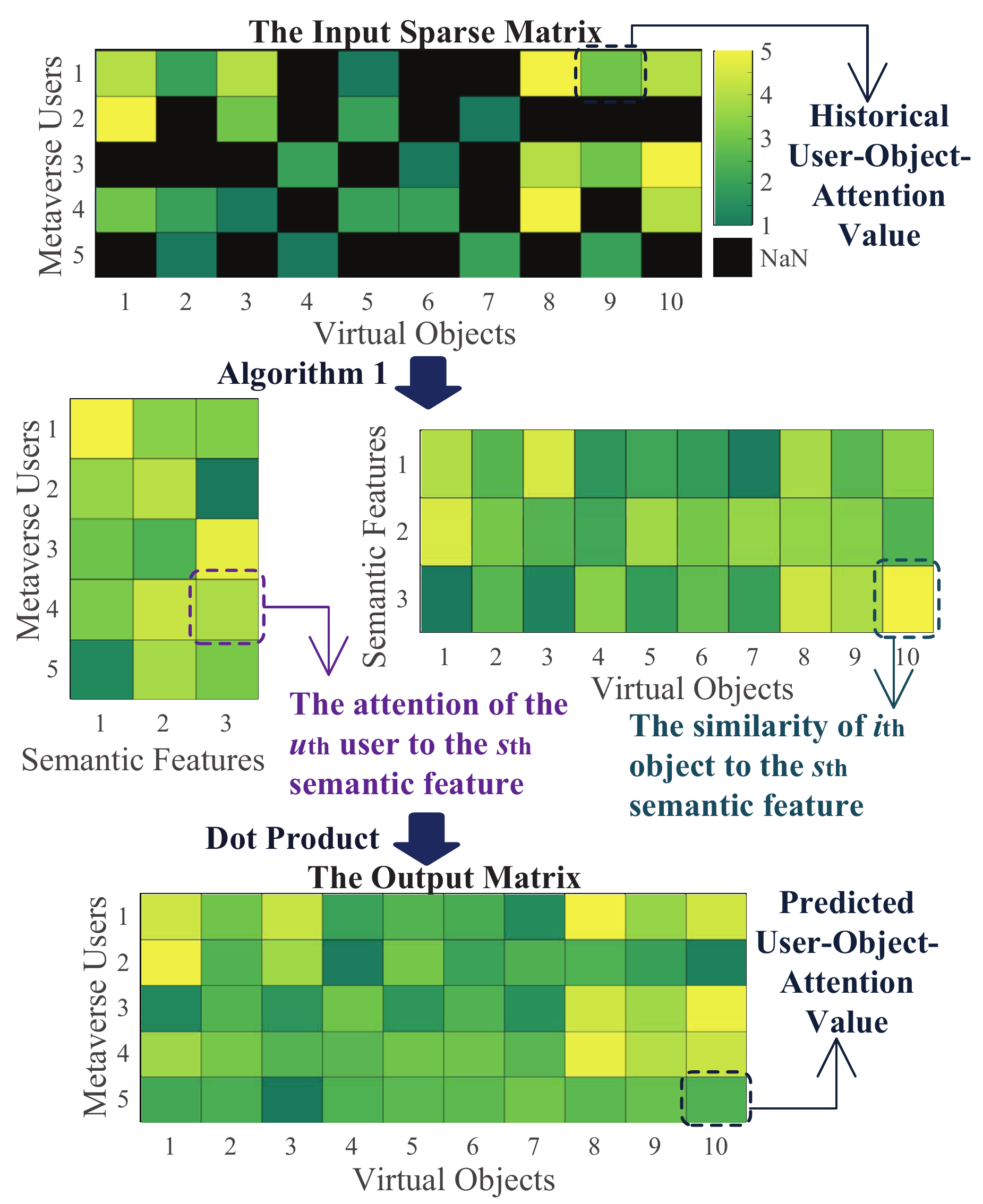}%
\caption{The working principle of attention-aware user-object-attention values prediction.}
\label{demo}
\end{figure}
Because Metaverse users have different attention to different virtual objects, and labels of objects are available to the MSP, we consider the virtual objects as {\it observed features} of users. The user's historical attention records to {\it observed features} are sparse~\cite{koren2009matrix}, because the user has unseen objects. Clearly, we can cluster a group of objects with semantic relevance into a ``topic''. The ``topics'' can be regarded as {\it semantic features}. For example, in the UOAL dataset, we have $96$ virtual objects such as ``mirror'', ``ball'', ``box'', and ``poster'', the {\it semantic features} could be ``toys'', ``tools'', etc. Although we cannot divide objects into independent ``topics'', it is reasonable to obtain a matrix that shows the similarities of each object to all semantic features. For example, ``mirror'' and ``box'' have high similarity to ``tools'' and low similarity to ``toys'', ``ball'' is the reverse, and the similarities of ``poster'' to both ``tools'' and ``toys'' are low, but might be relatively higher to ``toys''. Note that we do not need to consider the real meaning of {\it semantic features}, we only need to input the number of {\it semantic features}. As shown in Fig.~\ref{demo}, if we can estimate accurately the attention of users to {\it semantic features}, we can obtain the complete user-object-attention matrix by dot production~\cite{koren2009matrix}.

Based on the above discussion and motivated by the matrix factorization method~\cite{koren2009matrix}, we map Metaverse users and virtual objects into a joint semantic space that has {\small $S$} features. Then, the user-object-attention values can be modeled as inner products in the semantic space. Specifically, let {\small $N_U$} and {\small $N_O$} denote the number of users and objects in a user-object-attention matrix {\small ${\bm{A}}\in \mathbb{R}^{{N_U}\times {N_O}}$}, respectively. The semantic factor matrix for users and objects then can be expressed as {\small ${\bm{M}}\in \mathbb{R}^{N_U\times S}$} and {\small ${\bm{N}}\in \mathbb{R}^{N_O\times S}$}, respectively. The set of objects that have historical attention records with the $u_{\rm th}$ user is denoted by {\small ${\bm{\mathcal{A}}}_u$}. Similarly, for the $i_{\rm th}$ object, the set of Metaverse users that have attention records is denoted by {\small ${\bm{\mathcal{A}}}_i$}. The semantic feature vectors for the $u_{\rm th}$ user and the $i_{\rm th}$ virtual object are denoted by ${\bm m}_u\in \mathbb{R}^{S\times 1}$ and ${\bm n}_i \in \mathbb{R}^{S\times 1}$, respectively. Thus, the element $a_{ui}$ in ${\bm A}$ can be estimated as~\cite{koren2009matrix}
\begin{equation}\label{eq1}
{\hat a_{ui}}  = {\bm{m}_u}^{\rm T}{\bm{n}_i}.
\end{equation}
A weighted regression function-based method was proposed in \cite{hu2008collaborative} to learn the unknown model parameters. Similarly, we can obtain the predicted element ${{\hat a}_{ui}}$ by minimizing the following function:
\begin{align}\label{regreJ}
J \!= \!\sum\limits_{u = 1}^{{N_U}} {\sum\limits_{i = 1}^{N_O} {{w_{ui}}} } {\left( {{a_{ui}}\! - \!{{\hat a}_{ui}}} \right)^2} 
\!+\! \lambda \!\left( \!{\sum\limits_{u = 1}^{N_U} {{{\left\| {{{\bm{m}}_u}} \right\|}^2}}  \!\!+ \!\!\sum\limits_{i = 1}^{N_O} {{{\left\| {{{\bm{n}}_i}} \right\|}^2}} } \!\right)\!,
\end{align}
where $w_{ui}$ is the weight of $a_{ui}$, and $\lambda$ denotes the regularization strength to prevent over-fitting.

To optimize the regression model, i.e., \eqref{regreJ}, we use the Alternating Least Square (ALS) method~\cite{zachariah2012alternating}. With respect to Metaverse user semantic vector ${\bm m}_u$, minimizing $J$ in \eqref{regreJ} is equivalent to minimizing~\cite{hu2008collaborative}
\begin{equation}\label{recomj}
{J_u} = {\left\| {{{\bm{W}}^u}\left( {{{\bm{a}}_u} - {\bm{N}}{{\bm{m}}_u}} \right)} \right\|^2} + \lambda {\left\| {{{\bm{m}}_u}} \right\|^2},
\end{equation}
where {\small ${{\bm{W}}^u}$} is an $N_O \times N_O$ diagonal matrix, where the $\left(i_{\rm th},i_{\rm th}\right)$ element in {\small ${{\bm{W}}^u}$} is ${w}_{ui}$. To minimize \eqref{recomj}, we let the first-order derivative of ${J_u}$ be equal to $0$, and obtain
\begin{equation}\label{gfvaegv}
\frac{\partial {J_u}}{\partial {{\bm{m}}_u}}
= 2{{\bm{N}}^{\rm T}}{{\bm{W}}^u}{\bm{N}}{{\bm{m}}_u} - 2{{\bm{N}}^{\rm T}}{{\bm{W}}^u}{{\bm{a}}_u} + 2\lambda {{\bm{m}}_u} = 0.
\end{equation}
Thus, the corresponding ${{\bm{m}}_u}$ can be solved from \eqref{gfvaegv} as
\begin{equation}\label{gfvaegvgfvaegv}
{{\bm{m}}_u} = {\left( {{{\bm{N}}^{\rm T}}{{\bm{W}}^u}{\bm{N}} + \lambda {\bm{I}}} \right)^{ - 1}}{{\bm{N}}^{\rm T}}{{\bm{W}}^u}{{\bm{a}}_u},
\end{equation}
where {\small ${\bm{I}}$} is the identity matrix. Next, we can fix ${{\bm{m}}_u}$ and solve ${\bm n}_i$ by following the same process.

However, updating semantic vectors constrains the performance of ALS. To increase the processing speed for supporting online prediction, one solution is to optimize parameters at the element level. By optimizing each coordinate of the semantic vector while leaving the others fixed~\cite{rendle2011fast}, the computing speed can be greatly increased. Let $m_{uf}$ and $n_{if}$ denote the $f_{\rm th}$ element in ${{\bm{m}}_u}$ and ${{\bm{n}}_i}$, respectively. With respect to $m_{uf}$, we obtain the derivative of \eqref{recomj} as
\begin{equation}
\frac{\partial {J_u}}{\partial {{m_{uf}}}}\!\! =\! 2\!\left(\!\! {{m_{uf}}\!\sum\limits_{i = 1}^N {{w_{ui}}} n_{if}^2 \!-\!\! \sum\limits_{i = 1}^N \!{\left(\! {{a_{ui}}\! -\! \hat a_{ui}^f} \!\right)} {w_{ui}}{n_{if}} \!\!+\! \lambda {m_{uf}}} \!\!\!\right)\!\!,
\end{equation}
where $ \hat a_{ui}^f$ denotes the predicted element without the component of latent factor $f$, which means $ \hat a_{ui}^f = {{\hat a}_{ui}} - {m_{uf}}{n_{if}} $. Letting {\small $\frac{\partial {J_u}}{\partial {{m_{uf}}}} = 0$}, we obtain the solution of $m_{uf}$ as
\begin{equation}\label{muf}
{m_{uf}} = \frac{{\sum\limits_{i = 1}^N {\left( {{a_{ui}} - \hat a_{ui}^f} \right)} {w_{ui}}{n_{if}}}}{{\sum\limits_{i = 1}^N {{w_{ui}}} n_{if}^2 + \lambda }}.
\end{equation}
Then, ${n_{if}}$ can be derived by following the same method as
\begin{equation}\label{nif}
{n_{if}} = \frac{{\sum\limits_{u = 1}^M {\left( {{a_{ui}} - \hat a_{ui}^f} \right)} {w_{ui}}{m_{uf}}}}{{\sum\limits_{i = 1}^M {{w_{ui}}} m_{uf}^2 + \lambda }}.
\end{equation}
The detail is shown in {\textbf{Algorithm~\ref{semantiuca}}}.
\begin{algorithm}[t]
{\small \caption{The user-object-attention values prediction algorithm.} 
\label{semantiuca}
\hspace*{0.02in} {\bf Input:}
The sparse matrix with empty user-object-attention values, i.e., {\small ${\bm A}$}, such as shown in Fig. \ref{InputF}, the semantic factor $S$, regularization strength $\lambda$.\\
\hspace*{0.02in} {\bf Output:}
The predicted user-object-attention matrix without empty elements.
\begin{algorithmic}[1]
\State Initialize randomly {\small ${\bm M}$} and {\small ${\bm N}$}
\State Obtain the set of user-object pairs whose values are non-zero from {\small ${\bm A}$} as {\small ${\bm{\mathcal A}}$}
\For{Every $\left(u,i\right) \in {\bm{\mathcal A}}$ }
\State Calculate \eqref{eq1} to obtain ${\hat a_{ui}}$
\EndFor
\While{Estimation error is larger than the threshold}
\State {\it{( Update the semantic vector of users )}}
\For{$u = 1, \ldots, N_U$}
\For{$f = 1, \ldots, S$}
\For{$i \in {\bm{\mathcal A}}_u$}
\State $ \hat a_{ui}^f  \leftarrow  {{\hat a}_{ui}} - {m_{uf}}{n_{if}} $
\EndFor
\State Calculate \eqref{muf} to obtain $m_{uf}$
\For{$i \in {\bm{\mathcal A}}_u$}
\State $ \hat a_{ui}^f  \leftarrow  {{\hat a}_{ui}} + {m_{uf}}{n_{if}} $
\EndFor
\EndFor
\EndFor
\State {\it{( Update the semantic vector of objects )}}
\For{$i = 1, \ldots, N_O$}
\For{$f = 1, \ldots, S$}
\For{$u \in {\bm{\mathcal A}}_i$}
\State $ \hat a_{ui}^f  \leftarrow  {{\hat a}_{ui}} - {m_{uf}}{n_{if}} $
\EndFor
\State Calculate \eqref{nif} to obtain $n_{if}$
\For{$u \in {\bm{\mathcal A}}_i$}
\State $ \hat a_{ui}^f  \leftarrow  {{\hat a}_{ui}} + {m_{uf}}{n_{if}} $
\EndFor
\EndFor
\EndFor
\EndWhile
\State \Return {\small ${\bm M}\cdot{\bm N}$}.
\end{algorithmic}}
\end{algorithm}

We analyze the complexity of {\textbf{Algorithm~\ref{semantiuca}}}. If we use the ALS method with the help of \eqref{gfvaegv} and \eqref{gfvaegvgfvaegv}, time complexity is high because of the Matrix inversion operation, which can be expressed as {\small $\mathcal{O}\left( S^3\right) $}~\cite{he2016fast}. Considering that updating one semantic vector of Metaverse user has the complexity of {\small $\mathcal{O}\left( S^3 + N_O S^2 \right) $}, we can express the overall time complexity that updates all parameters once as {\small $\mathcal{O}\left( \left( N_O+N_U\right) S^3 + N_UN_O S^2 \right) $}. Fortunately, by using \eqref{muf} and \eqref{nif}, we can avoid the matrix inversion operation and reach the complexity of {\small $\mathcal{O}\left( N_UN_O S^2 \right) $} for one iteration. Furthermore, if the CBS can pre-compute $ {\hat r}_{ui}$, $ {\hat r}_{ui}^f$ can be obtained with the complexity of {\small $\mathcal{O}\left(1\right) $}. Thus, the complexity of {\textbf{Algorithm~\ref{semantiuca}}} is further reduced to {\small $\mathcal{O}\left( N_UN_O S \right) $}, which can support online user-object-attention predictions.
Note that although many more sophisticated machine learning-based methods can be applied to obtain potentially more accurate predictions, our method is simple and fast, and can support achieving QoE close to the upper limit. Specifically, in Section~\ref{nums}, we show that the achieved QoE is only $2\%$ lower than that when all user-object-attention values are perfectly known.

\subsubsection{Optimal Rendering Capacity Allocation}\label{attentioncon}
Let {\small $ {P_k^{\left( R \right)}} $} denote the total rendering capacity allocated to one virtual scenario chosen by the $k_{\rm th}$ user. The sum of resolutions set for each virtual object cannot be larger than {\small $ {P_k^{\left( R \right)}} $}. We then design the optimal rendering capacity allocation scheme by solving the maximization problem as follows:
\begin{equation}\label{filling}
\begin{array}{*{20}{c}}
{\mathop {\max }\limits_{P_{1,k}^{\left( R \right)}, \ldots ,P_{{N_{Ok}},k}^{\left( R \right)}} }&{\sum\limits_{n = 1}^{{N_{Ok}}} {{K_{n,k}}\ln \left( {\frac{{P_{n,k}^{\left( R \right)}}}{{P_{\rm th}^{\left( R \right)}}}} \right)} }\\
{\rm{s.t.}}&{\left\{ {\begin{array}{*{20}{c}}
{P_{n,k}^{\left( R \right)} > {P_{\rm th}^{\left( R \right)}},\forall n,}\\
{\sum\limits_{n = 1}^{{N_{Ok}}} {P_{n,k}^{\left( R \right)}}  \le P_k^{\left( R \right)}.}
\end{array}} \right.}
\end{array}
\end{equation}
When one user chooses a Metaverse service, the user-object-attention values $K_{n,k}$ can be predicted with the help of {\textbf{Algorithm~\ref{semantiuca}}}. Then we derive the optimal rendering capacity allocation scheme.
\begin{prop}\label{renderinglem}
To maximize the MI, the rendering capacity for each object in Metaverse xURLLC services is allocated as
\begin{equation}
{P_{n,k}^{\left( R \right)}}^* = \max \left\{ {{K_{n,k}}\frac{1}{{{\mu ^*}}},P_{\rm th}^{\left( R \right)}} \right\},
\end{equation}
where ${\mu ^*}$ is obtained by solving the following function
\begin{equation}\label{fgeagea}
\sum\limits_{n = 1}^{{N_{Ok}}} {\max \left\{ {{K_{n,k}}\frac{1}{{{\mu ^*}}},P_{\rm th}^{\left( R \right)}} \right\}}  = P_k^{\left( R \right)}.
\end{equation}
\end{prop}
\begin{IEEEproof}
Please refer to Appendix \ref{renderingapp}.
\end{IEEEproof}
With the help of Proposition~\ref{renderinglem}, we propose the optimal rendering capacity allocation algorithm as shown in Algorithm \ref{Fills}. Since the number of iterations of {\textbf{Algorithm~\ref{Fills}}} is equal to the number of objects that are allocated with the minimal rendering capacity, i.e., {\small $P_{\rm th}^{\left( R\right) }$}, Algorithm \ref{Fills} can converge efficiently and output the optimal rendering capacity allocation scheme.

\begin{algorithm}[t]
{\small \caption{The algorithm for allocating the rendering capacity with the help of predicted user-object-attention values.} 
\label{Fills}
\hspace*{0.02in} {\bf Input:}
The predicted user-object-attention values from {\textbf{Algorithm~\ref{semantiuca}}}, \small{$K_{n,k}$}, and the total rendering capacity, {\small $P_k^{\left( R\right) }$}.\\
\hspace*{0.02in} {\bf Output:}
The optimal rendering capacity allocation scheme, {\small ${P_{1,k}^{\left( R \right)}, \ldots ,P_{{N_{Ok}},k}^{\left( R \right)}}$}.
\begin{algorithmic}[1]
\State Initialize one temporary variable $j=1$ and two temporary lists {\small $T_1$} and {\small $T_2$}
\State Start by assuming {\small ${P_{n,k}^{\left( R \right)}}^* = {{K_{n,k}}\frac{1}{{{\mu ^*}}}} $} for all $n$
\State Calculate {\small ${\mu ^*} \leftarrow  \sum\limits_{n = 1}^{{N_{Ok}}} {{K_{n,k}}} / P_k^{\left(R\right) }$}, and then {\small ${P_{n,k}^{\left( R \right)}}^* \leftarrow {K_{n,k}}/{\mu ^*}$} for every $n$
\While{ the minimum of {\small ${P_{n,k}^{\left( R \right)}}^*$} $<$ {\small $P_{\rm th}^{\left( R\right) }$}}
\State Record {\small $T_1\left[ j \right]$ $\leftarrow$} the object number of minimal {\small ${P_{n,k}^{\left( R \right)}}^*$}
\State Record {\small $T_2\left[ j \right]$ $\leftarrow$} the user-object-attention value for the {\small $t_1\left[ j \right]_{\rm th}$} object
\State Re-calculate {\small ${\mu ^*}$} and {\small ${P_{n,k}^{\left( R \right)}}^*$}:
\State \qquad {\small $ {\mu ^*} \leftarrow  \left( \sum\limits_{n = 1}^{{N_{Ok}}} {{K_{n,k}}} -\sum {{T_2}} \right) / \left(P_k^{\left(R\right)}-j\times P_{\rm th}^{\left( R\right) } \right) $}
\State \qquad {\small ${P_{n,k}^{\left( R \right)}}^* \leftarrow {K_{n,k}}/{\mu ^*}$} for every $n$
\For{${\rm temp} = 1:j$}
\State Allocate {\small $P_{\rm th}^{\left( R\right) }$} to the {\small $t_1\left[ {\rm temp} \right]_{\rm th}$} object:
\State \qquad {\small $n \leftarrow t_1\left[ {\rm temp} \right]_{\rm th}$}, {\small ${P_{n,k}^{\left( R \right)}}^* \leftarrow P_{\rm th}^{\left( R\right) }$}
\EndFor
\State $j \leftarrow j+1$
\EndWhile
\State \Return The optimal rendering capacity allocation scheme {\small ${{P_{1,k}^{\left( R \right)}}^*, \ldots ,{P_{{N_{Ok}},k}^{\left( R \right)}}^*}$}.
\end{algorithmic}}
\end{algorithm}

{\textbf{Algorithms~\ref{semantiuca} and \ref{Fills}}} form the attention-aware rendering capacity allocation algorithm. Thus, the InP can make better use of resources in the wireless MIMO network to provide Metaverse xURLLC services. 

\section{QoE Analysis and Contract Solution}\label{SR5}
In addition to the personalized attention-aware rendering power allocation scheme for xURLLC discussed, we have to derive the expressions for URLLC KPIs, i.e., downlink data rate and uplink BEP, under the considered wireless MIMO network to obtain the closed-form MI expression. Then, we analyze the convexity and discuss the contract solution.

\subsection{URLLC KPIs Analysis}\label{KPSI}
To obtain the analytical expression and investigate the convexity of MI, we derive the closed-form expressions of {\small ${R_k^{\left( D \right)}}$} and {\small $P_k^{\left( U \right)}$}, respectively.
\subsubsection{Downlink Data Rate}\label{ssa}
Following the analysis in Section \ref{MTN}, we can derive the closed-form expression of data rate {\small ${R_k^{\left( D \right)}}$} as follows:
\begin{prop}\label{dataratelemma}
The data rate {\small ${R_k^{\left( D \right)}}$} can be expressed as~\footnote{To obtain the approximated achievable data rate for xURLLC packets, the short packet transmission can be used to achieve the low latency requirement, which can be addressed in our future work.}
\begin{equation}\label{datarate}
R_k^{\left( D \right)} =\! \frac{{B_k}{{\Gamma ^{ - 1}}\left( {{M_C}{N_Q}} \right)}}{{{\ln\left(2 \right) }\Gamma \! \left( {{M_C}{M_U}} \right)}}G_{3,3}^{3,2}\!\left(\! {{\Lambda _k}\left|\!\!\! {\begin{array}{*{20}{c}}
{1 - {M_C}{N_Q},0,1}\\
{{M_C}{M_U},0,0}
\end{array}} \right.} \!\!\!\right),
\end{equation}
where $B_k$ is the bandwidth allocated to the $k_{\rm th}$ user, $G \, \substack{ \cdot, \cdot \\ \cdot, \cdot}(\cdot)$ is the Meijer's $G$-function~\cite[eq. (9.301)]{gradshteyn2007} and $\Gamma \! \left( \cdot \right) $ is the Gamma function~\cite[eq. (8.310.1)]{gradshteyn2007}.
\end{prop}
\begin{IEEEproof}
Please refer to Appendix \ref{datarateapp}.
\end{IEEEproof}

{\color{black} \noindent{\textbf{Insights:}} From the derived date rate, i.e., \eqref{datarate}, we can obtain useful insights to guide the URLLC system design. Specifically, when co-channel interference is high in the wireless environment, i.e., $P_{k,q}\to 0$, we have $ {\Lambda _k} \to 0$. Equation \eqref{datarate} can be further simplified by calculating the residue at the nearest pole to the Mellin-Barnes integral contour of the Meijer's $G$-function~\cite{gradshteyn2007}. Thus, the data rate can be approximated to
	\begin{equation}
	{\hat R}_k^{\left( D \right)} \approx {M_C}^3{M_U}{N_Q}\frac{{{B_k}{\zeta _k}D_k^{ - {\alpha _k}}P_k^{\left( D \right)}{\mu _k}}}{{\ln \left( 2 \right){P_{k,q}}{\mu _{k,q}}}}.
\end{equation}
We can observe that increasing the number of RS antennas can better improve the anti-interference capability of URLLC system compared with increasing the number of CBS antennas.
}

\subsubsection{Uplink Bit Error Probability}
We consider that the frequency division multiplexing is used in the downlink and uplink between CBSs and RSs. For the uplinks, let {\small $ {{P_p}^{\left( U \right)}} $} denote the co-channel interference, {\small $ {{P_k}^{\left( U \right)}} $} denote the uplink transmit power,  $ {{\zeta ^{\left( U \right)}}} $ can be calculated by \eqref{agfeaga}, $ {{\mu _k}^{\left( U \right)}} $ and $ {{\mu _p}^{\left( U \right)}} $ denote the channel coefficients for the $k_{\rm th}$ user's links and corresponding interference links, respectively.

The average BEP, $E_k^{\left(U \right)}$, under a variety of modulation formats is given by~\cite{tse2005fundamentals}
\begin{equation}
E_k^{\left(U \right)} = \int_0^\infty  {\frac{{\Gamma \! \left( {{\tau _2},{\tau _1}\gamma } \right)}}{{2\Gamma \! \left( {{\tau _2}} \right)}}{f_{{\gamma _{k}^{\left(U \right) }}}}\left( \gamma  \right)d\gamma },
\end{equation}
where {\small ${f_{{\gamma _{k}^{\left(U \right) }}}}$} is the PDF expression of the uplink SIR, i.e., {\small $\gamma _{k}^{\left(U \right) }$}, ${\tau _1} $ and ${\tau _2}$ are modulation-specific parameters for several modulation and detection scheme combinations, $ {{{\Gamma \! \left( {{\tau _2},{\tau _1}\gamma } \right)}}/{{2\Gamma \! \left( {{\tau _2}} \right)}}} $ represents the conditional bit-error probability, and $\Gamma \! \left( { \cdot , \cdot } \right)$ denotes the upper incomplete Gamma function~\cite[eq. (8.350.2)]{gradshteyn2007}. When the orthogonal coherent binary frequency-shift keying (BFSK) scheme is used, we set $\left\{ {{\tau _1} = 0.5,{\tau _2} = 0.5} \right\}$. When the antipodal coherent binary phase-shift keying (BPSK) scheme is applied, we have $\left\{ {{\tau _1} = 1,{\tau _2} = 0.5} \right\}$. For the orthogonal non-coherent BFSK scheme, we have $\left\{ {{\tau _1} = 0.5,{\tau _2} = 1} \right\}$. For the antipodal differentially coherent BPSK (DPSK) scheme, we set $\left\{ {{\tau _1} = 1,{\tau _2} = 1} \right\}$.
\begin{prop}\label{berlemma}
The uplink BEP is given by
\begin{align}\label{berfinal}
E_k^{\left(U \right)}  \!=& \frac{{{\Gamma ^{ - 1}}\!\left( {{M_U}{N_Q}} \right)}}{{2\Gamma \! \left( {{\tau _2}} \right)\!\Gamma \! \left( {{M_C}{M_U}} \right)}}G_{4,1}^{1,3}\!\!\left(\!\! {\frac{{{\Lambda _k^{\left( U\right) }}}}{{{\tau _1}}}\!\left|\!\!\! {\begin{array}{*{20}{c}}
{1 \!-\! {M_U}{N_Q},1,1 \!-\! {\tau _2}}\\
{{M_C}{M_U},0}
\end{array}} \right.} \!\!\!\!\!\right)\!,
\end{align}
where
\begin{equation}\label{delku}
{\Lambda _k}^{\left( U \right)} = \frac{{{P_p}^{\left( U \right)}{\mu _p}^{\left( U \right)}}}{{{M_U}{\zeta ^{\left( U \right)}}{P_k}^{\left( U \right)}{\mu _k}^{\left( U \right)}}}.
\end{equation}
\end{prop}
\begin{IEEEproof}
Please refer to Appendix \ref{berapp}.
\end{IEEEproof}

{\color{black} \noindent{\textbf{Insights:}} Similar to our analysis in Section~\ref{ssa}, when the co-channel interference is high, we consider that ${\Lambda _k^{\left( U \right)}} $ in~\eqref{berfinal} approaches 0, which leads to $E_k^{\left(U \right)}$ converging to 0.5. Considering that high transmit power is typically used in URLLC to achieve low latency communication~\cite{she2021tutorial}, we consider the case when $P_k^{\left( U\right)} \to \infty$. According to the properties of the Meijer's $G$-function~\cite{gradshteyn2007}, we have
\begin{equation}
	\hat E_k^{\left( U \right)} \approx \frac{{\Gamma \left( {{M_C}{M_U} + {\tau _2}} \right){\Gamma ^{ - 1}}\left( {{\tau _2}} \right)}}{{2{M_C}{M_U}B\left( {{M_U}{N_Q},{M_C}{M_U}} \right)}}{\left( {\frac{{\Lambda _k^{\left( U \right)}}}{{{\tau _1}}}} \right)^{{M_C}{M_U}}}.
\end{equation}
We can observe that the decrease in BEP due to the increase in transmit power is proportional to the product of the numbers of RS and CBS antennas.
}

\subsection{Convexity Analysis}\label{convex}
Considering that {\small $\mathcal{M}_k$} appears in the utility functions of both the MSP and the InP, we study the convexity of {\small $\mathcal{M}_k$} to several resources, i.e., downlink transmit power {\small $P_k^{\left(D\right) }$}, downlink bandwidth {\small $B_k$}, uplink transmit power {\small $P_k^{\left(U\right) }$}, and rendering capacity {\small $P_{k}^{\left( R\right) }$}, with the help of derived equations in Propositions~\ref{dataratelemma} and \ref{berlemma}.
Because {\small $ {{\Lambda _k}} $} in \eqref{delk} is a function of {\small $P_k^{\left(D\right) }$} and {\small $ {{\Lambda _k^{\left(U\right) }}} $} in \eqref{delku} is a function of {\small $P_k^{\left(U\right) }$}, we first derive Lemma~\ref{delklemma} that is useful for the analysis of {\small $\mathcal{M}_k$}.
\begin{lemma}\label{delklemma}
Let {\small $x \triangleq {P_k^{\left( D \right)}}$}, {\small $y \triangleq {P_k^{\left( U \right)}}$} and $\forall \theta \in \left[ {0,1} \right] $ denote a real number. Then, we have {\small ${\left( {{\Lambda _k}\left( x \right)} \right)^s}$} $\left( -1<s<0\right) $ and {\small $ {\left( {\Lambda _k^{\left(U\right) }}{\left( y \right)}\right) }^t $} $\left( t>0\right) $ which satisfy the following inequalities:
\begin{equation}\label{delkuse}
{\left( {{\Lambda _k}\left( {\theta {x_1} \!+\! \left( {1 - \theta } \right){x_2}} \right)} \right)^s} \!\ge\! \theta \Lambda _k^s\left( {{x_1}} \right){{ + }}\left( {1 - \theta } \right)\Lambda _k^s\left( {{x_2}} \right)\!,
\end{equation}
and
\begin{equation}\label{delkuuse}
{\left(\! {\Lambda _k^{\left( U \right)}\left( {\theta {y_1} \!+ \!\left( {1 - \theta } \right){y_2}} \right)} \!\right)^t} \!\le\! \theta \Lambda _k^t\left( {{y_1}} \right)\!{{ + }}\!\left( {1 - \theta } \right)\Lambda _k^t\left( {{y_2}} \right)\!,
\end{equation}
respectively.
\end{lemma}
\begin{IEEEproof}
Please refer to Appendix \ref{applem}.
\end{IEEEproof}
Using Lemma~\ref{delklemma}, we can analyze the convexity of {\small $\mathcal{M}_k$}.
\begin{prop}\label{MIconv}
The MI of the $k_{\rm th}$ user, {\small $\mathcal{M}_k$}, is linear to the downlink bandwidth {\small $B_k$}, and concave in the downlink transmit power {\small $P_k^{\left(D\right) }$}, the uplink transmit power {\small $P_k^{\left(U\right) }$}, and the rendering capacity {\small $P_k^{\left(R\right) }$}.
\end{prop}
\begin{IEEEproof}
Please refer to Appendix \ref{MIconvapp}.
\end{IEEEproof}
With the expressions for the downlink data rate and the uplink BEP and the convexity analysis, we can obtain the closed-form MI expression. Thus, the optimal contract can be solved numerically.

\subsection{Contract Solution}
We first consider the IC constraint. The closed-form expression of MI can be obtained by substituting \eqref{datarate} and \eqref{berfinal} into \eqref{mi}. The InP's utility is concave in {\small ${{\bm{\Theta }}}$}, since the MI is concave in the resources as we discussed in Section~\ref{KPSI}.For a given contract {\small $ \left\{ {{F_s},{u_M}} \right\} $}, with the help of {\textbf{Algorithms~\ref{semantiuca}}} and \ref{Fills}, we can use convex optimization tools to obtain the optimal resource allocation scheme and corresponding $\mathcal{M}_i$. Accordingly, we substitute the IR constraint with the optimal amount of resource and simplify the MSP's maximization problem. The optimal contract, {\small $ \left\{ {{F_s}^*,{u_M}^*} \right\} $}, can be also solved using the convex method. We show the MI values obtained under the optimal resource allocation schemes for different contract designs in Fig.~\ref{NewModel} (Part IV). Detailed discussion is given in Section~\ref{nums}.

\section{Numerical Results}\label{nums}
In this section, we first introduce the UOAL dataset~\cite{du2022exploring}, and then present analytical results to illustrate the proposed Metaverse xURLLC service framework. System-wide simulation is left for our future research direction.

\subsection{Metaverse Dataset Preparation}\label{SectionData}
\begin{figure*}[t]
\centering
\includegraphics[width=0.85\textwidth]{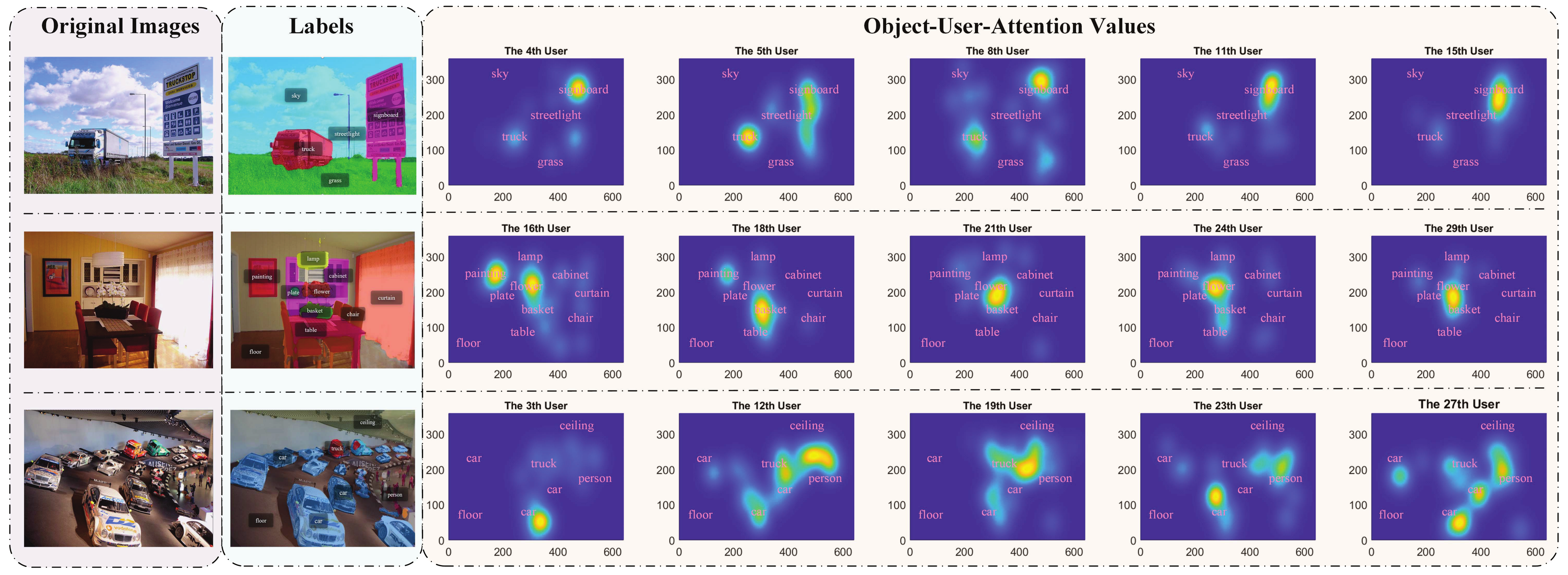}
\caption{Some examples for object labels and the user-object-attention values in UOAL~\cite{du2022exploring}.} 
\label{examples}
\end{figure*}
\begin{figure*}[t]
\centering
\includegraphics[height=0.3\textwidth,width=0.93\textwidth]{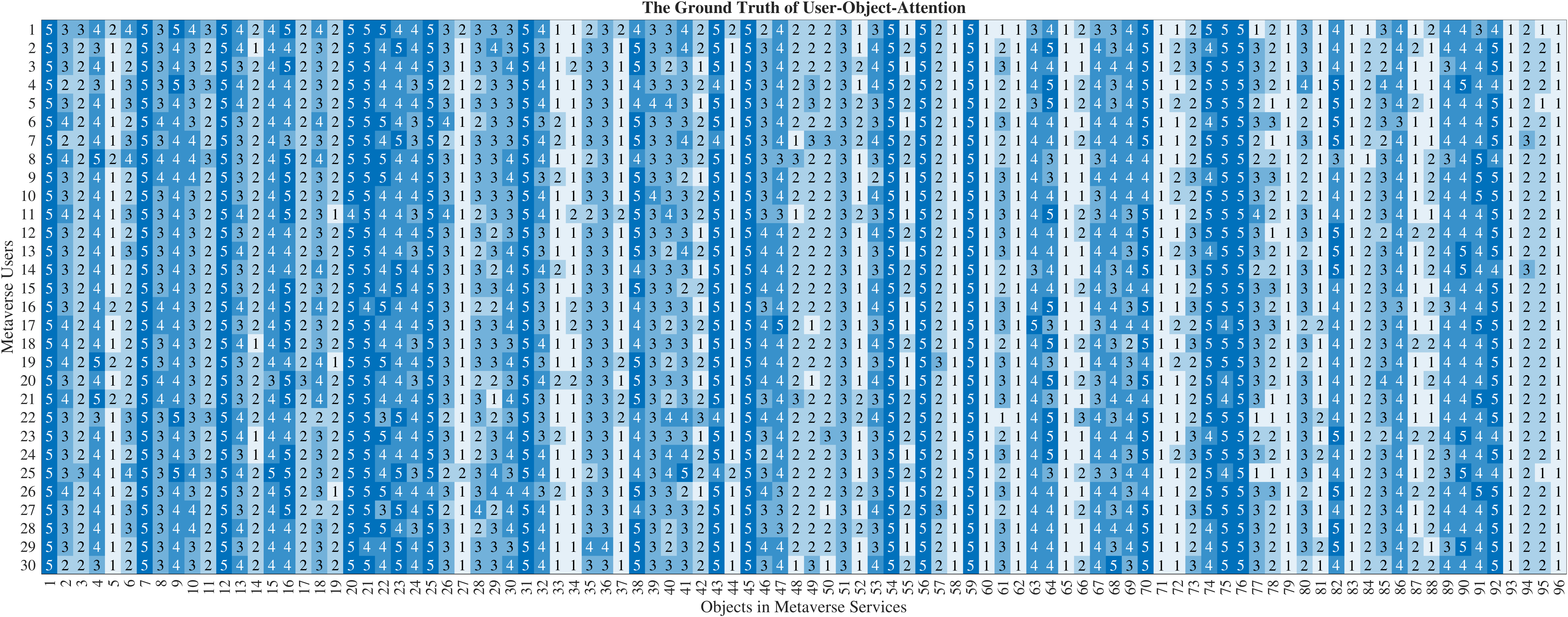}%
\caption{The ground truth of the user-object-attention records, which is obtained by letting every user see every images in UOAL.}
\label{GTUOA}
\end{figure*}
\begin{figure*}[t]
\centering
\includegraphics[height=0.3\textwidth,width=0.9\textwidth]{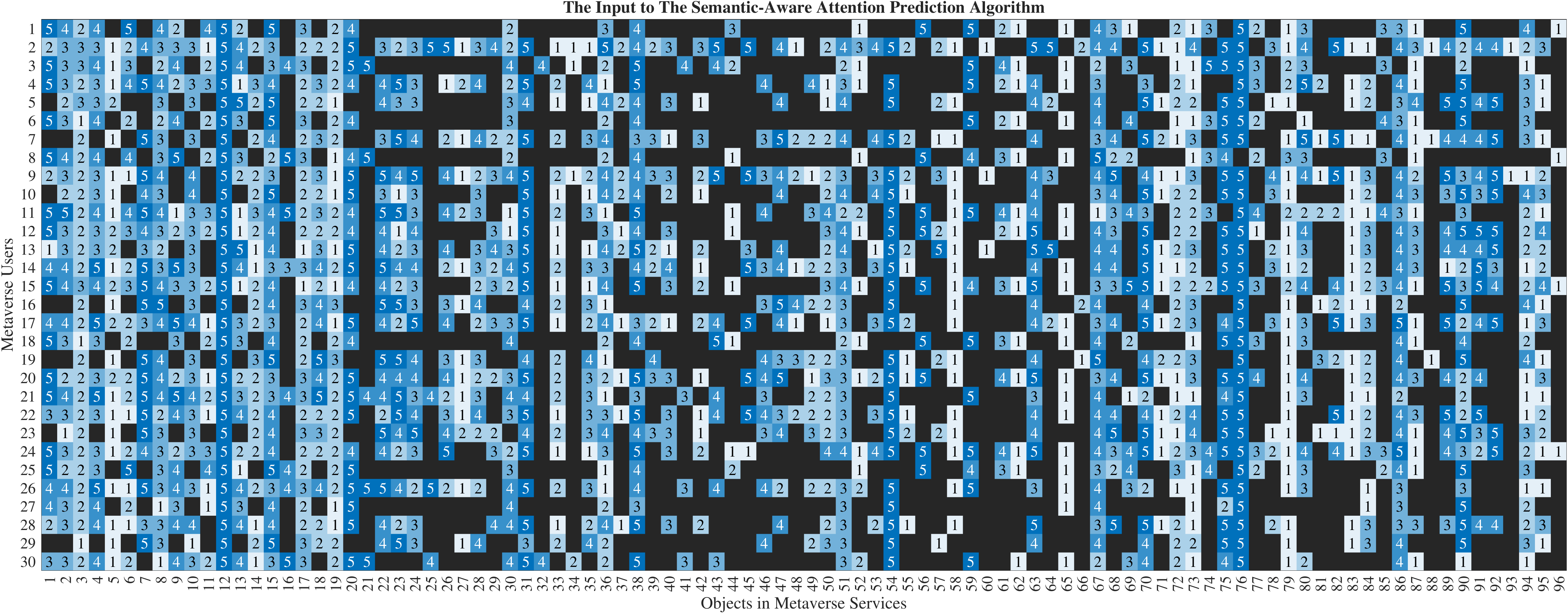}%
\caption{The sparse user-object-attention records obtained by randomly selecting the virtual scenarios options for each user, and randomly reserving the images as the freely-selected travel path for each user.}
\label{InputF}
\end{figure*}

In Metaverse context, we consider a virtual travel scenario, where the user clearly cannot simultaneously see all objects to be displayed, e.g., in VR, and the user's attention is unevenly distributed among the objects that are seen at the same time. This process is similar to that the users view a subset of images. Different users' attention to different objects contained in the images is different because of the users' personal interest. Therefore, we use the UOAL dataset~\cite{du2022exploring}, which is composed of $1,000$ images, $96$ different types of objects, $30$ users' attention values to every object. Specifically, some examples for object labels and the user-object-attention values in UOAL are shown in Fig. \ref{examples}. If we let all users see all the images in UOAL, we can obtain the ground truth of the user-object-attention records as shown in Fig. \ref{GTUOA}, which is not given in~\cite{du2022exploring}. To simulate a Metaverse xURLLC service, we generate sparse attention records between users and objects. All the images are divided into small groups, where each group represents one Metaverse virtual scenario option. By letting users select randomly the options and see a random number of images, e.g., due to different virtual travel routes in the Metaverse xURLLC services, we obtain the sparse user-object-attention records as shown in Fig. \ref{InputF} that is also the input to {\textbf{Algorithm~\ref{semantiuca}}}. The process of using UOAL to obtain the sparse records is given in Appendix \ref{Datasetapp}.

\subsection{Performance Analysis}
To answer the three questions we presented in Section~\ref{Intro}, we show the performance of the proposed schemes, including the attention-aware resource allocation scheme, i.e., user-object-attention values prediction in {\textbf{Algorithm~\ref{semantiuca}}} and optimal rendering capacity allocation in {\textbf{Algorithm~\ref{Fills}}}, and optimal contract design.
\begin{figure*}[t]
\centering
\includegraphics[height=0.3\textwidth,width=0.9\textwidth]{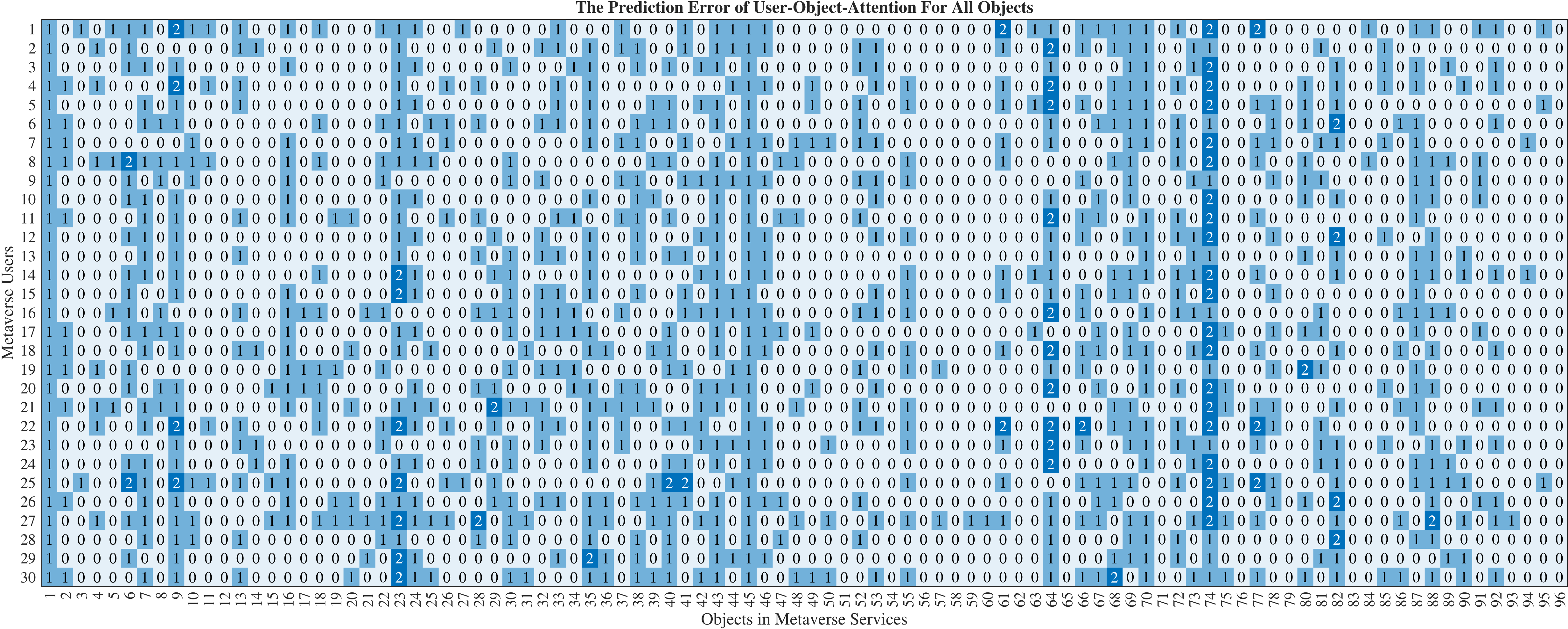}%
\caption{The prediction error of user-object-attention values for all objects.}
\label{PredAll}
\end{figure*}
\subsubsection{The accuracy of the attention-aware user-object-attention values prediction {\textbf{Algorithm~\ref{semantiuca}}} (for {\textbf{Q3}})}: To show the accuracy of {\textbf{Algorithm~\ref{semantiuca}}}, we first randomly generate the user-object-attention sparse matrix as shown in Fig.~\ref{InputF}, where $41.8\%$ of the elements are empty, as the input to {\textbf{Algorithm~\ref{semantiuca}}}. Figure~\ref{PredAll} shows the user-object-attention prediction error values, i.e., the absolute values of the difference between the predicted results and ground truth (Fig. \ref{GTUOA}). In Fig. \ref{PredAll}, we observe that $67.2\%$ of the prediction error values are $0$, $30.7\%$ of the error values are $1$, and only $2.12\%$ of the error values are $2$.
If we consider only the values that are unknown in the input matrix, we can observe that $62.8\%$ of the error values are $0$, $34.1\%$ of the error values are $1$, and $3.16\%$ of the error values are $2$.
Thus, we conclude that {\textbf{Algorithm~\ref{semantiuca}}} can help to predict the user-object-attention values accurately. In the following discussion, we show that the predicted values can be used to achieve QoE close to the theoretical upper bound.

\subsubsection{The effectiveness of the optimal rendering capacity allocation {\textbf{Algorithm~\ref{Fills}}} (for {\textbf{Q3}})}: Figure~\ref{NewModel} (Part III) illustrates the MI of $30$ users under three different allocation schemes, i.e., random, uniform (conventional URLLC without considering the attention differences among users), attention-aware rendering capacity allocation (xURLLC), and the upper-bound obtained from the ground truth. The resolution {\small $\left( {\rm K}\right)$} is used as the measure of rendering capacity\footnote{Here we define that 1 ${\rm K}$ resolution refers to $960\times 480$ pixel resolution \cite{webhua}. It is suggested that the resolution should be above 12 ${\rm K}$ to obtain the ideal-experience~\cite{webhua}.}. We consider that the minimal rendering capacity threshold for one virtual object, i.e., {\small $P_{\rm th}^{\left( R\right) }$}, is $15$ {\small ${\rm K}$}. The total rendering capacity for the $i_{\rm th}$ user is ${\small N_{Oi}}\times 20$ {\small ${\rm K}$}, where {\small $N_{Oi}$} is the number of virtual objects. We consider that the MSP has the sparse user-object-attention values matrix, i.e., Fig. \ref{InputF}. The Metaverse scenario options are then selected randomly by users according to Algorithm~\ref{algorithm1}, and then different resource allocation schemes are used. Specifically, the random and uniform rendering capacity allocation scheme distribute the rendering capacity randomly and uniformly, respectively, while ensuring that each object is assigned with the minimum threshold {\small $P_{\rm th}^{\left( R\right) }$}. For the attention-aware scheme, the MSP predicts the user-object-attention values with the help of {\textbf{Algorithm~\ref{semantiuca}}} to obtain {\small ${K_{n,k}}$}. Thus, the rendering capacity is allocated with the help of {\textbf{Algorithm~\ref{Fills}}}. The upper-bound is obtained with the help of ground truth user-object-attention values and {\textbf{Algorithm~\ref{Fills}}}. From Fig.~\ref{NewModel} (Part III), we can observe that the xURLLC attention-aware rendering capacity allocation scheme can achieve a maximum of $25.5\%$, a minimum of $6.26\%$ and an average of $20.1\%$ MI improvement compared to the URLLC uniform rendering capacity allocation scheme. Moreover, the theoretical upper-bound has only a $2\%$ QoE improvement compared to the attention-aware scheme. This verifies the effectiveness of our proposed optimal rendering capacity allocation {\textbf{Algorithm~\ref{Fills}}}

{\small \begin{table}
\caption{Network Parameters for $3$ Metaverse users.}
\label{parameters}
\centering
\renewcommand{\arraystretch}{1}
{\small\begin{tabular}{m{3.2cm}<{\centering}|m{1.2cm}<{\centering}|m{1.25cm}<{\centering}|m{1.25cm}<{\centering}}
\hline
\textbf{Parameters} & $1^{\rm st}$ User & $2^{\rm nd}$ User& $3^{\rm rd}$ User \\
\hline
Number of antennas in CBS, $M_C$ & \multicolumn{3}{c}{6} \\
\hline
Number of antennas in RS, $M_U$ & \multicolumn{2}{c|}{3}&7 \\
\hline
Number of co-channel interference, $N_Q$ & \multicolumn{3}{c}{3} \\
\hline
Power of interference links, ${{P_{k,p}}}$ & \multicolumn{2}{c|}{5 ${\rm dBW}$} & 1 ${\rm dBW}$ \\
\hline
Channel coefficient of interference links, ${{\mu _{k,p}}}$ & -3 ${\rm dB}$ & -1 ${\rm dB}$& -3 ${\rm dB}$ \\
\hline
Channel coefficient of data links, ${{\mu _{k}}}$ & -1 ${\rm dB}$ & -2 ${\rm dB}$& -1 ${\rm dB}$ \\
\hline
Distance between CBS and RS, ${{D _{k}}}$ & 10 ${\rm m}$ & 6 ${\rm m}$ & 10 ${\rm m}$ \\
\hline
Path loss exponent, ${{\alpha_{k}}}$ & \multicolumn{3}{c}{2} \\
\hline
\end{tabular}}
\end{table}}
\begin{figure}[t]
\centering
\includegraphics[width=0.34\textwidth]{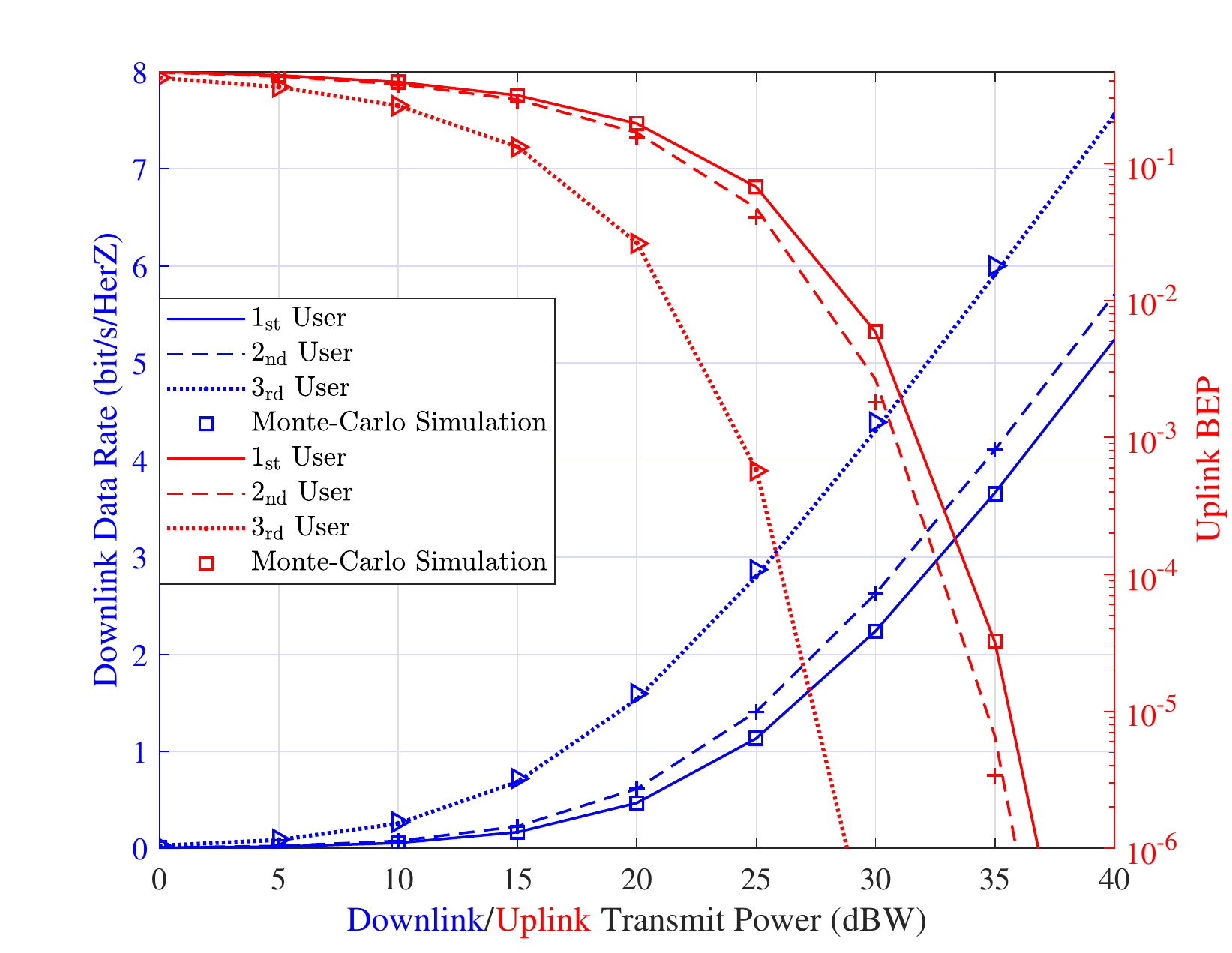}
\caption{Conventional URLLC KPIs in the Meta-Immersion expression, i.e., downlink data rate and uplink BEP, for three users versus the downlink and uplink transmit power, respectively.}
\label{wireless}
\end{figure}
\begin{figure}[t]
\centering
\includegraphics[width=0.34\textwidth]{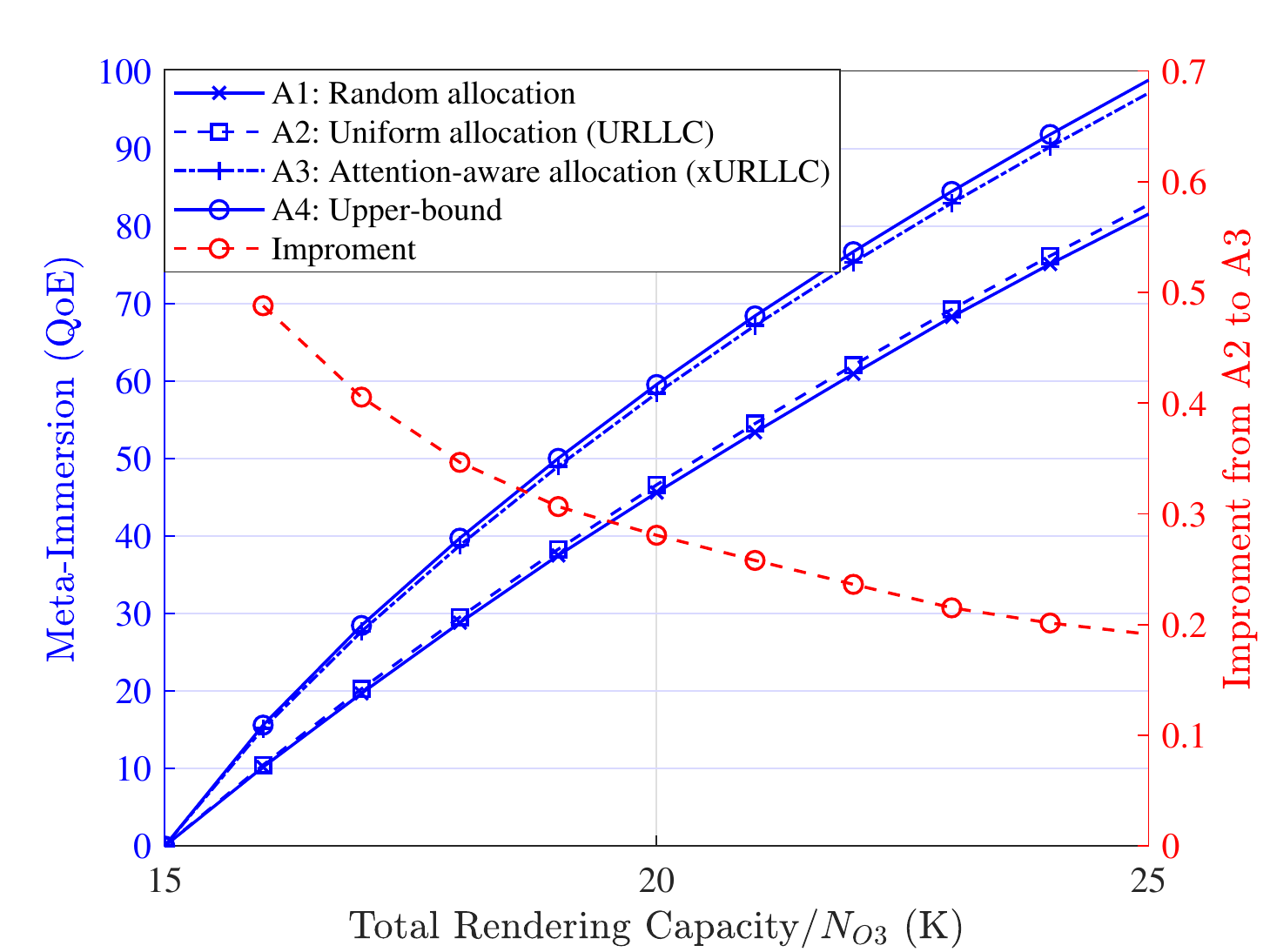}
\caption{Meta-Immersion of the $3^{\rm rd}$ user versus the total rendering capacity, under three different resource allocation schemes, and the upper-bound obtained using the ground truth user-object-attention values.}
\label{renderdel}
\end{figure}

\subsubsection{The superiority of the xURLLC over conventional URLLC (for {\textbf{Q2}} and {\textbf{Q3}})}: Considering that the QoE expression contains the objective KPIs, i.e., downlink data rate and uplink BEP, which are affected by the wireless transmission in the MIMO network, we consider three Metaverse users with the parameters that are given in Table~\ref{parameters}. The analytical results of downlink data rate and uplink BEP are obtained from \eqref{datarate} and \eqref{berfinal}, respectively. Perfect agreement is observed between analytical results and Monte Carlo simulations, thus validating our derivations. Furthermore, we can observe the impacts of different network parameters on users' wireless connections. Specifically, Table~\ref{parameters} shows that the $2^{\rm nd}$ user is in a better quality wireless environment than the $1^{\rm st}$ user. {\color{black} As show in Fig.~\ref{wireless}, compared with the $1^{\rm st}$ user, the $2^{\rm nd}$ user has a higher downlink data rate and a lower uplink BEP under the same transmit power. Hardware facilities also affect the wireless connections. The RS serving the $3^{\rm th}$ user is equipped with the maximum number of antennas, thus allowing the $3^{\rm th}$ user to achieve the best performance.}
Fortunately, in the MIMO wireless network, satisfactory URLLC KPIs can always be achieved when the transmit power is high.
However, conventional URLLC that does not utilize the uneven user attention in novel Metaverse services to achieve personalized rendering capability allocation would bring users lower QoE values than xURLLC. To show this, we focus on the $3^{\rm rd}$ user in Fig.~\ref{NewModel} (Part III). For the virtual scenario option selected randomly by the $3^{\rm rd}$ user, there are $N_{O3} = 56$ objects. {\color{black} Figure \ref{renderdel} plots the MI of the $3^{\rm rd}$ user versus the total rendering capacity/$N_{O3}$ under three different resource allocation schemes and the upper-bound obtained from the ground truth. An interesting insight is that the attention-aware scheme brings a higher percentage of improvement compared to the uniform allocation scheme when the total resources are constrained, i.e., when the total rendering capacity is small. This shows that the attention-aware scheme can improve resource utilization efficiency and thus bring a better Metaverse experience to users.}
\begin{figure}[t]
\centering
\includegraphics[width=0.36\textwidth]{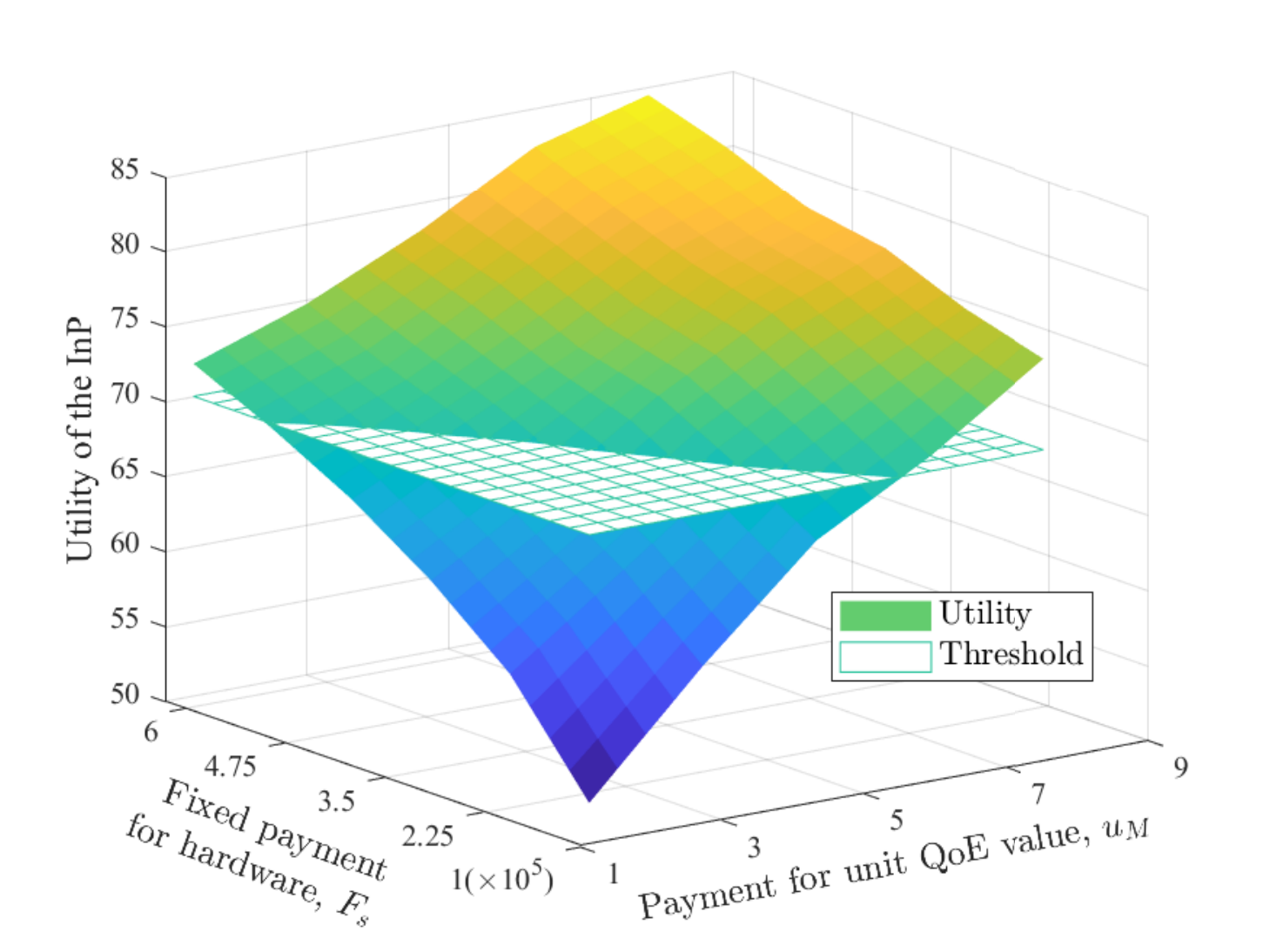}%
\caption{The utility of the InP versus the payment from the MSP to the InP for unit QoE value, $u_M$, and the fixed payment from the MSP for using the hardware infrastructures, $F_s$.}
\label{UINP}
\end{figure}
\begin{figure}[t]
\centering
\includegraphics[width=0.36\textwidth]{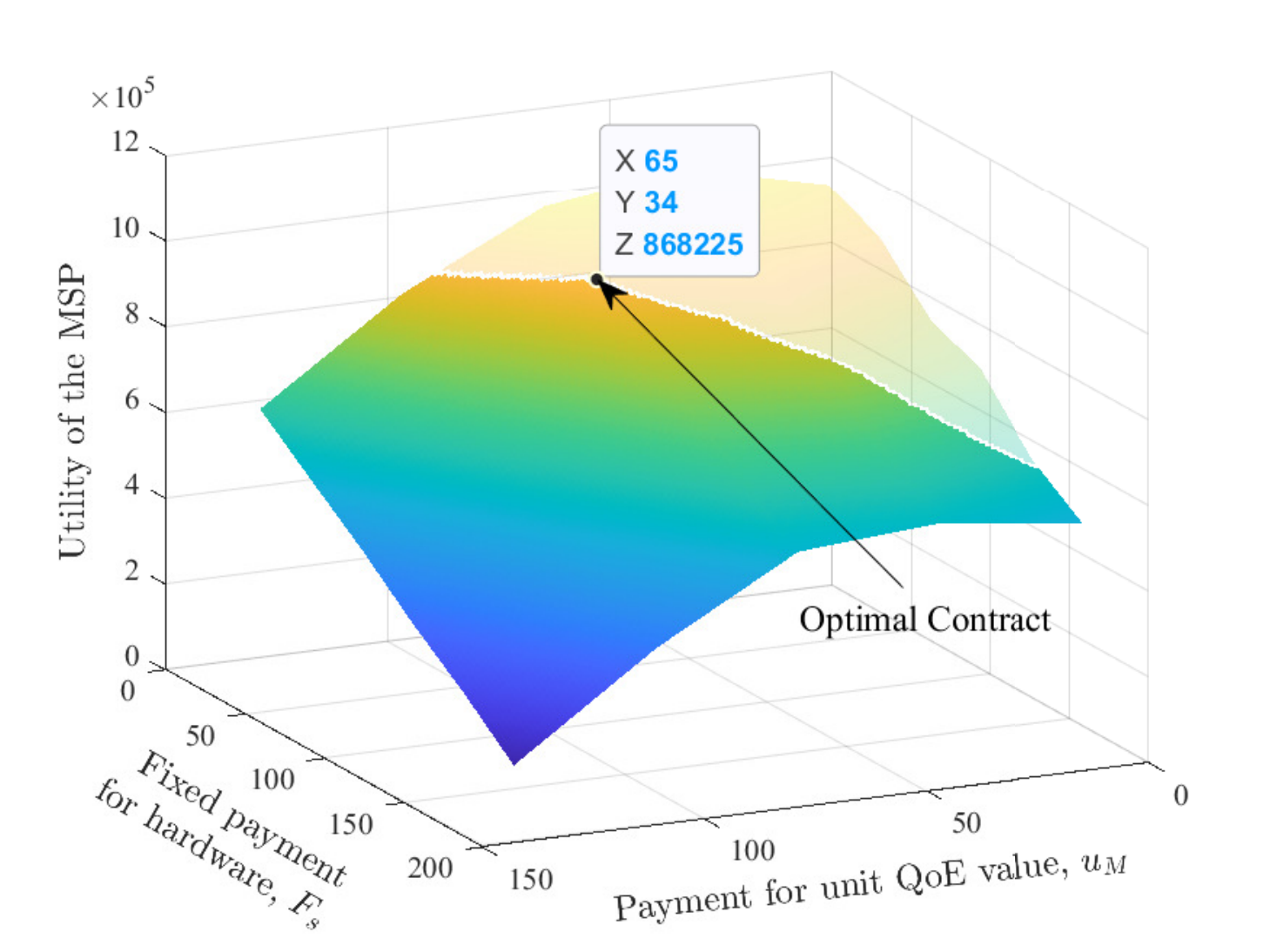}%
\caption{The utility of the MSP versus the payment from the MSP to the InP for unit QoE value, $u_M$, and the fixed payment from the MSP for using the hardware infrastructures, $F_s$.}
\label{UMSP}
\end{figure}

\subsubsection{The effectiveness of the optimal contract design (for {\textbf{Q1}})}: We then study the optimal contract design problem in the Metaverse xURLLC service market. Figs~\ref{NewModel} (Part IV), \ref{UINP}~\ref{UMSP} depict the MI, utilities of the InP and the MSP versus the payment from the MSP to the InP for unit QoE value $u_M$ and the fixed payment from the MSP for using the hardware infrastructures {\small $F_s$}, respectively, with the network parameters for $3$ Metaverse users as shown in Table~\ref{parameters}, {\small $U_{\rm th}^{\rm InP} = 70$}, $\tau=0.8$, {\small ${R_{k,{\rm max}}^{\left( D\right) }}=42$} ${\rm Mbit/s}$, {\small ${R_{k,{\rm min}}^{\left( D\right) }}=10$} ${\rm Mbit/s}$, {\small ${E_{k,{\rm min}}^{\left(U\right) }}=10^{-2}$}, {\small ${E_{k,{\rm min}}^{\left(U\right) }}=10^{-8}$}, and the unit prices for rendering capacity (per {\small ${\rm K}$}), downlink transmit power (per {\small ${\rm kW}$}), downlink bandwidth (per {\small ${\rm MHz}$}), uplink power (per {\small ${\rm kW}$}), are denoted by ${\bm \Theta} = \{5,3,2,4\}$.
Specifically, Fig. \ref{UINP} shows the IC constraint. For a given contract {\small $ \left\{ {{F_s},{u_M}} \right\} $}, the InP optimizes its utility function based on the price of various resources and the contract. An interesting insight is that the optimal resource allocation scheme that leads to the maximal MI is only related to ${u_M}$ rather than {\small ${F_s}$}. The reason is that {\small ${F_s}$} is an additive term in the utility of the InP, \eqref{reutilityInP}, which is independent of MI. However, it is clear that {\small ${F_s}$} impacts the value of the InP's utility function and the IR constraint. From Fig. \ref{UINP}, we can observe that the increase of both ${u_M}$ and {\small ${F_s}$} leads to an increase in utility. Because of the IR constraint, the utility of the InP must be larger than a threshold, {\small $U_{\rm th}^{\rm InP}$}, which makes the contracts with small values of {\small${F_s}$} and {\small${u_M}$} not available. Thus, the value of {\small ${F_s}$} cannot be zero. The optimal contract obtained by maximizing the utility of the MSP under IC and IR constraints is shown in Fig. \ref{UMSP}. We can observe that the utility of the MSP under the optimal contract decreases as the {\small $U_{\rm th}^{\rm InP}$} required by the InP increases. However, the optimal contract scheme, i.e., the marked point in Fig. \ref{UMSP}, can always achieve the highest utility values for both the InP and the MSP, compared with other feasible contract schemes. This verifies the effectiveness of the optimal contract design.

\section{{\color{black} Conclusion and Future Directions}}\label{S7}
To deploy Metaverse xURLLC services in the wireless MIMO network, we proposed a contract design framework between the MSP and the InP. The utility of the MSP was maximized while ensuring the QoE of users, i.e., KPI of the xURLLC service, and the incentives of the InP. To consider both the objective KPIs and the subjective feelings of Metaverse users, we proposed a novel metric, i.e., MI, to define formally the QoE for next-generation Internet services such as Metaverse xURLLC. By analyzing the objective network performance indicators and subject user attention values, the closed-form expression of MI was derived. Considering that the historical user-object-attention records in the Metaverse service are sparse, we designed the attention-aware rendering capacity allocation algorithm that can predict the attention values first and then allocate resources optimally. Using the UOAL dataset, we verified that the MI can be increased averagely by $20.1\%$ by the xURLLC attention-aware allocation scheme compared to the conventional URLLC uniform allocation scheme.

{\color{black} We list potential future research directions as follows:
	\begin{itemize}
		\item {\textit{Cold-start in Metaverse services:}} In this paper, we use sparse user-object interaction records to predict user attention. However, the cold-start problem exists when the users are new. To enable personalized resource allocation scheme for any user without collecting extra user information beforehand, cold-start problem has to be addressed. A possible solution is to use graph neural networks to perform interest prediction from the user's current Metaverse services~\cite{du2022exploring}.
		\item {\textit{Attention-aware cross-layer design:}} We focus on using the user attention mechanism to guide the upper-layer resource allocation scheme design. As emerging physical layer access technologies, e.g., extremely large-scale MIMO~[54], are increasingly being applied, cross-layer design schemes can be investigated to achieve URLLC and further improve the QoE. For example, the accuracy of the beamforming scheme can be dynamically adjusted according to the Metaverse service content.
		\item {\textit{Secure Metaverse services:}} We derive the downlink data rate and the uplink BEP for the MIMO technique to provide Metaverse services. However, data in the wireless environment can be easily eavesdropped, thus posing a privacy risk to users. To achieve secure wireless network-enabled Metaverse services, physical layer security, covert communication, and encryption techniques can be applied~\cite{chorti2022context}.
\end{itemize}}


\begin{appendices}
\section{Dataset Preparation}\label{Datasetapp}
\renewcommand{\theequation}{A-\arabic{equation}}
\setcounter{equation}{0}
The algorithm that is used to obtain the sparse records are given in Algorithm \ref{algorithm1} in Python style.
\begin{algorithm}[h]
\caption{Generating sparse user-object-attention records for the $k_{\rm th}$ user in Metaverse xURLLC service.}\label{algorithm1}
{\small 
\begin{lstlisting}[language=python]
		# G, E, L: grouped images, eye-movement records 
		#  and segmentation labels, all in array format
		# a1: random int range [2, 4]
		# a2: random float range [0.3, 0.7]
		
		# randomly select a1 groups
		G_k, E_k, L_k = selet_data(G, E, L, a1)
		# randomly reserve a2 of data
		G_k, E_k, L_k = reserve_data(G_k, E_k, L_k, a2)
		
		scores = []
		for j in object_list:
		# count the frequency of object j 
		c_j = count_frequency(j, G_k)
		
		# find the positions of j on segmentations
		pos_j = numpy.where(L_k==j)
		
		# compute attention score of j
		s_j = E_k[pos_j].sum()/c_j
		
		scores.append(s_j)
		
		# split scores into 5 levels
		levels = numpy.array_split(scores.sort(), 5)
		# map raw values to 5 levels
		scores_k = map_level(scores, levels)
\end{lstlisting}}
\end{algorithm}

\section{Proof of Proposition \ref{dataratelemma}}\label{datarateapp}
\renewcommand{\theequation}{B-\arabic{equation}}
\setcounter{equation}{0}
The downlink data rate per Hertz is defined as
\nobreak\begin{equation}\label{aefjeal}
R_k^{\left( B\right) } \triangleq R_k^{\left( D\right) }/B_k = \int_0^\infty  {\log_2 \left( {1 + \gamma } \right){f_{\gamma_k} }\left( \gamma  \right){\rm d}\gamma}.
\end{equation}
Substituting \eqref{PDF} into \eqref{aefjeal}, we can obtain
\nobreak\begin{align}\label{sdafkh}
R_k^{\left( B \right)} =& \frac{{{\Lambda _k}^{{M_C}{M_U}}}}{{B\!\left( {{M_C}{M_U},{M_C}{N_Q}} \right)}}
\notag\\&\times
\int_0^\infty  {\frac{{{{\log }_2}\!\left( {1 \!+\! x} \right){x^{{M_C}{M_U} - 1}}}}{{{{\left( {1 \!+\! x{\Lambda _k}} \right)}^{{M_C}{M_U} + {M_C}{N_Q}}}}}{\rm{d}}x}.
\end{align}
With the help of~\cite[eq. (01.04.07.0003.01)]{web}, the logarithmic function in \eqref{sdafkh} can be re-written as
\begin{equation}
{\log _2}\left( {1 + x} \right) = \frac{1}{{2\pi i}}\int_{{\cal L}_1} {\frac{{\Gamma \! \left( {s + 1} \right){\Gamma ^2}\left( { - s} \right){x^{ - s}}}}{{\Gamma \! \left( {1 - s} \right)}{\ln 2}}} {\rm{d}}s,
\end{equation}
where the integration path of $\mathcal{L}_1$ goes from $s -i\infty $ to $s+i\infty $, $s$ is a real number, $-1<s<0$ and $i =\sqrt{-1}$. Then, the $R_k^{\left( B \right)}$ can be expressed as
\begin{equation}\label{eaga4eg}
R_k^{\left( B \right)} \!=\! \frac{{{\Lambda _k}^{{M_C}{M_U}}}}{{B \! \left( {{M_C}{M_U},{M_C}{N_Q}} \right)}}\frac{1}{{2\pi i}}\int_{{\cal L}_1} {\frac{{\Gamma \! \left( {s\! +\! 1} \right){\Gamma ^2}\left( { - s} \right)}}{{\Gamma \! \left( {1 - s} \right)}}} {I_C}{\rm{d}}s,
\end{equation}
where
\begin{equation}
{I_C} = \int_0^\infty  {\frac{{{x^{{M_C}{M_U} - 1 - s}}}}{{{{\left( {1 + x{\Lambda _k}} \right)}^{{M_C}{M_U} + {M_C}{N_Q}}}}}{\rm{d}}x}.
\end{equation}
According to~\cite[eq. (3.194.3)]{gradshteyn2007}, the $I_C$ can be solved as~\cite{du2021millimeter}
\begin{equation}\label{akvjg}
{I_C}= {\Lambda _k}^{s - {M_C}{M_U}}\frac{{\Gamma \! \left( {{M_C}{M_U} - s} \right)\Gamma\!\left( {s + {M_C}{N_Q}} \right)}}{{\Gamma\!\left( {{M_C}{M_U} + {M_C}{N_Q}} \right)}}.
\end{equation}
By combining \eqref{akvjg} and \eqref{eaga4eg}, we can obtain
\begin{align}\label{meilinR}
& R_k^{\left( B \right)} =  \frac{1}{{\Gamma \! \left( {{M_C}{M_U}} \right)\Gamma \! \left( {{M_C}{N_Q}} \right)}}\frac{1}{{2\pi i}}
\notag\\&\times
\int_{{\cal L}_1} {\frac{{{\Gamma ^2}\left( { - s} \right)\Gamma\!\left( {s + {M_C}{N_Q}} \right)\Gamma \! \left( {s + 1} \right)}}{{{\Gamma ^{ - 1}}\left( {{M_C}{M_U} - s} \right)\Gamma \! \left( {1 - s} \right)}}} {\Lambda_k^s}{\rm{d}}s.
\end{align}
With the help of~\cite[eq. (9.301)]{gradshteyn2007} and \eqref{aefjeal}, we derive the closed-form {\small $R_k^{\left( D \right)}$} as \eqref{datarate}, which completes the proof.

\section{Proof of Proposition \ref{berlemma}}\label{berapp}
\renewcommand{\theequation}{C-\arabic{equation}}
\setcounter{equation}{0}
Using the definition of Gamma function~\cite[eq. (8.350)]{gradshteyn2007}, we can re-write $E_k^{\left(U \right)}$ as
\begin{equation}\label{ekudef}
E_k^{\left(U \right)} = \frac{{{\tau _1}^{{\tau _2}}}}{{2\Gamma \! \left( {{\tau _2}} \right)}}\int_0^\infty  {{x^{{\tau _2} - 1}}{e^{ - {\tau _1}x}}{F_{{\gamma _{k}^{\left(U \right) }}}}\left( x \right){\rm{d}}\gamma },
\end{equation}
where {\small ${F_{{\gamma _{k}^{\left(U \right) }}}}$} is the CDF expression for the uplink SIR, $\gamma_k^{\left(U \right) }$. We first derive {\small ${F_{{\gamma _{k}^{\left(U \right) }}}}$}. Using the definition of CDF, we have
\begin{equation}\label{gaehsrh}
{F_{{\gamma _k^{\left(U \right) }}}}\left( \gamma  \right){\rm{ = }}\int_{\rm{0}}^\gamma  {{f_{{\gamma _k}}^{\left(U \right) }}\left( x \right){\rm{d}}x}.
\end{equation}
Replacing {\small $ {{M_C}{N_Q}} $} with {\small $ {{M_U}{N_Q}} $} and substituting \eqref{PDF} into \eqref{gaehsrh}, we then solve the integration part with the help of~\cite[eq. (3.194)]{gradshteyn2007} and obtain
\begin{align}
&	{F_{{\gamma _k^{\left(U \right) }}}}\left( \gamma  \right){{ = }}\frac{{{\Lambda_k^{\left(U\right) }}^{{M_C}{M_U}}}}{{B \! \left( {{M_C}{M_U},{M_U}{N_Q}} \right)}}\frac{{{\gamma ^{{M_C}{M_U}}}}}{{{M_C}{M_U}}}
\notag\\&\times\!\!
{}_{\rm{2}}{F_{\rm{1}}}\!\!\left(\!{{M_C}{M_U} \!+\! {M_U}{N_Q},{M_C}{M_U};\!1\! + \!{M_C}{M_U};\! \frac{{\Lambda _k^{\left(U\right) }}}{-\gamma} } \!\right)\!,
\end{align}
where ${}_2{F_1}\left( { \cdot , \cdot ; \cdot ; \cdot } \right)$ is the Gauss hypergeometric function \cite[eq. (9.111)]{gradshteyn2007}. With the help of~\cite[9.113]{gradshteyn2007} and~\cite[8.331.1]{gradshteyn2007}, the CDF can be expressed as
\begin{align}\label{CDF}
&	{F_{{\gamma _k^{\left(U \right) }}}}\left( \gamma  \right){{ = }}\frac{{\rm{1}}}{{\Gamma \! \left( {{M_C}{M_U}} \right)\Gamma \! \left( {{M_U}{N_Q}} \right)}}\frac{1}{{2\pi i}}
\notag\\&\times
\int_{{\cal L}_2} {\frac{{\Gamma \! \left( {{M_U}{N_Q} + t} \right)\Gamma \! \left( t \right)}}{{{\Gamma ^{ - 1}}\left( {{M_C}{M_U} - t} \right)\Gamma \! \left( {1 + t} \right)}}{{\left( {{\Lambda _k^{\left(U\right) }}\gamma } \right)}^t}{\rm{d}}t},
\end{align}
where the integration path of $\mathcal{L}_2$ goes from $t -i\infty $ to $t+i\infty $, $t$ is a real number, and $t>0$. Substituting \eqref{CDF} into \eqref{ekudef}, the BEP can be written as
\begin{align}\label{faeg3}
&	E_k^{\left(U \right)} = \frac{{{\tau _1}^{{\tau _2}}{\Gamma ^{ - 1}}\left( {{M_U}{N_Q}} \right)}}{{2\Gamma \! \left( {{\tau _2}} \right)\Gamma \! \left( {{M_C}{M_U}} \right)}}\frac{1}{{2\pi i}}
\notag\\&\times
\int_{{\cal L}_2} {\frac{{\Gamma \! \left( {{M_U}{N_Q} + t} \right)\Gamma \! \left( t \right)}}{{{\Gamma ^{ - 1}}\left( {{M_C}{M_U} - t} \right)\Gamma \! \left( {1 + t} \right)}}{\Lambda _k^{\left(U\right) }}^t{I_D}{\rm{d}}t},
\end{align}
where
\begin{equation}
{I_D}{{ = }}\int_0^\infty  {{x^{t + {\tau _2} - 1}}{e^{ - {\tau _1}x}}{\rm{d}}x}.
\end{equation}
With the help of~\cite[eq. (3.351.3)]{gradshteyn2007} and~\cite[eq. (8.339.1)]{gradshteyn2007}, $I_D$ can be solved as 
\begin{equation}\label{id}
{I_D}{{ = }}\Gamma \! \left( {t + {\tau _2}} \right){\tau _1}^{ - t - {\tau _2}}.
\end{equation}
Substituting $I_D$ into \eqref{faeg3} and using~\cite[eq. (9.301)]{gradshteyn2007}, we can derive \ref{berfinal} to complete the proof.

\section{Proof of Proposition \ref{renderinglem}}\label{renderingapp}
\renewcommand{\theequation}{D-\arabic{equation}}
\setcounter{equation}{0}
The objective function in \eqref{filling} is convex in the rendering capacity and the strong duality holds for the convex problem according to the Slater's condition~\cite{boyd2004convex}. The Lagrangian associated with problem \eqref{filling} is given by
\begin{align}
{F_{\cal L}}  \triangleq & {F_{\cal L}}\left( {P_{1,k}^{\left( R \right)}, \ldots ,P_{{N_{Ok}},k}^{\left( R \right)},{\lambda _1}, \ldots ,{\lambda _{{N_{Ok}}}},\mu } \right)
\notag\\
= & -\sum\limits_{n = 1}^{{N_{Ok}}} {{K_{n,k}}\ln \left( {\frac{{P_{n,k}^{\left( R \right)}}}{{P_{\rm th}^{\left( R \right)}}}} \right)}  - \sum\limits_{n = 1}^{{N_{Ok}}} {{\lambda _n}\left( {P_{n,k}^{\left( R \right)} - P_{\rm th}^{\left( R \right)}} \right)}
\notag\\&
- \mu \left( {P_k^{\left( R \right)} - \sum\limits_{n = 1}^{{N_{Ok}}} {P_{n,k}^{\left( R \right)}} } \right),
\end{align}
where ${\lambda _n}$ $\left(n=1,\ldots,N_{Ok} \right) $ and $\mu$ are the Lagrange multipliers. The Karush–Kuhn–Tucker (KKT) optimality conditions for the optimal solution is
\begin{equation}\label{kkt}
\left\{ \begin{array}{l}
-{K_{n,k}} + P{_{n,k}^{\left( R \right)} }\left( {\mu  - {\lambda _n}} \right) = 0,\\
P{_{n,k}^{\left( R \right)} } - P_{\rm th}^{\left( R \right)} \ge 0,\\
{\lambda _n}  \ge 0,\\
{\lambda _n} \left( {P{{_{n,k}^{\left( R \right)}} } - P_{\rm th}^{\left( R \right)}} \right) = 0,\\
P_k^{\left( R \right)} - \sum\limits_{n = 1}^{{N_{Ok}}} {P{{_{n,k}^{\left( R \right)}} }}  = 0,
\end{array} \right.
\end{equation}
which leads to
\begin{equation}\label{solu}
{\lambda _n} = -{K_{n,k}}\frac{1}{{P_{n,k}^{\left( R \right)}}} + \mu .
\end{equation}
By \eqref{kkt} and \eqref{solu}, we have two cases for solving optimal {\small $ {P{{_{n,k}^{\left( R \right)}} }} $} as follows:

{\textbf {\it Case 1:}} {\small $ {\lambda _n} > 0 \: \& \:  P_{n,k}^{\left( R \right)} = P_{\rm th}^{\left( R \right)} $}

By solving \eqref{solu}, we have $\mu  < \frac{{ - {K_{n,k}}}}{{P_{\rm th}^{\left( R \right)}}}$.

{\textbf {\it Case 2:}} {\small $ {\lambda _n} = 0 \: \& \:  P_{n,k}^{\left( R \right)} \ge P_{\rm th}^{\left( R \right)} $}

By solving \eqref{solu}, we have $P_{n,k}^{\left( R \right)} = {K_{n,k}}\frac{1}{\mu } \ge P_{\rm th}^{\left( R \right)}$.

Therefore, it can be inferred from the above two cases that
\begin{equation}
{P_{n,k}^{\left( R \right)}}^* = \max \left\{ {{K_{n,k}}\frac{1}{{{\mu ^*}}},P_{\rm th}^{\left( R \right)}} \right\},
\end{equation}
where $ {\mu ^*} $ can be obtained by solving
\begin{equation}
\sum\limits_{n = 1}^{{N_{Ok}}} {P{{_{n,k}^{\left( R \right)}}^*}}  = \sum\limits_{n = 1}^{{N_{Ok}}} {\max \left\{ {{K_{n,k}}\frac{1}{{{\mu ^*}}},P_{\rm th}^{\left( R \right)}} \right\}}  = P_k^{\left( R \right)}.
\end{equation}
Thus, we complete the proof.

\section{Proof of Lemma \ref{delklemma}}\label{applem}
\renewcommand{\theequation}{E-\arabic{equation}}
\setcounter{equation}{0}
We study the convexity of {\small $ {\left( {{\Lambda _k}\left( x \right)} \right)^s} $} $\left( -1<s<0\right) $ and {\small $ {\left( {{\Lambda _k}^{\left( U \right)}\left( y \right)} \right)^t} $} $\left( t>0\right) $. By performing the quadratic derivative to $x$ and $y$, we obtain
\begin{equation}
\frac{{{\partial ^2}{{\left( {{\Lambda _k}\left( x \right)} \right)}^s}}}{{\partial {x^2}}} = {\left( {\frac{{{P_p}{\mu _p}}}{{{M_U}\zeta {\mu _k}}}} \right)^s}\frac{{s\left( {s + 1} \right)}}{{{x^{s + 2}}}} < 0,
\end{equation}
and
\begin{equation}
\frac{{{\partial ^2}{{\left( {\Lambda _k^{\left( U \right)}\left( y \right)} \right)}^t}}}{{\partial {y^2}}} = {\left( {\frac{{{P_p}{\mu _p}}}{{{M_U}\zeta {\mu _k}}}} \right)^t}\frac{{t\left( {t + 1} \right)}}{{{x^{t + 2}}}} > 0,
\end{equation}
respectively.

Thus, {\small $ {\left( {{\Lambda _k}\left( x \right)} \right)^s} $} is concave and {\small $ {\left( {{\Lambda _k}^{\left( U \right)}\left( y \right)} \right)^t} $} is convex. With the help of the Jensen's inequality~\cite{dragomir1994some}, we can derive \eqref{delkuse} and \eqref{delkuuse} to complete the proof.

\section{Proof of Proposition \ref{MIconv}}\label{MIconvapp}
\renewcommand{\theequation}{F-\arabic{equation}}
\setcounter{equation}{0}
We first focus on {\small $P_k^{\left(D\right) }$}. Let {\small $x \triangleq P_k^{\left(D\right) }$} and $\forall \theta \in \left[ {0,1} \right] $ denote a real number. Using \eqref{meilinR} in Proposition~\ref{dataratelemma}, we have
\begin{align}
R_k^{\left( B \right)}&\left( {\theta {x_1} + \left( {1 - \theta } \right){x_2}} \right) = \frac{1}{{\Gamma\!\left( {{M_C}{M_U}} \right)\Gamma\!\left( {{M_C}{N_Q}} \right)}}\frac{1}{{2\pi i}}
\notag\\&\times
\int_{{L_1}} {\frac{{{\Gamma ^2}\left( { - s} \right)\Gamma\!\left( {s + {M_C}{N_Q}} \right)\Gamma\!\left( {s + 1} \right)}}{{{\Gamma ^{ - 1}}\left( {{M_C}{M_U} - s} \right)\Gamma\!\left( {1 - s} \right)}}}
\notag\\&\times
\Lambda _k^s\left( {\theta {x_1} + \left( {1 - \theta } \right){x_2}} \right){\rm{d}}s.
\end{align}
With the help of \eqref{delkuse} in Lemma~\ref{delklemma}, after some mathematical transformations, we obtain
\begin{equation}
R_k^{\left( B \right)}\!\left( {\theta {x_1}\! +\! \left( {1 \!- \!\theta } \right){x_2}} \right) \!\ge\! \theta R_k^{\left( B \right)}\!\left( {{x_1}} \right)\!{{ + }}\!\left( {1\! -\! \theta } \right)R_k^{\left( B \right)}\left( {{x_2}} \right)\!.
\end{equation}
According to the definition of convex function~\cite{boyd2004convex}, we conclude that {\small $R_k^{\left( B \right)}$} is concave in {\small $P_k^{\left(D\right) }$}. Considering the {\small $\mathcal{T}\left( \cdot\right)$} \eqref{calT} function in {\small $\mathcal{M}_k$} is the linear-fractional function that preserves convexity~\cite{boyd2004convex}, we can derive that {\small $\mathcal{M}_k$} is concave in {\small $P_k^{\left(D\right) }$}.

For downlink bandwidth {\small $B_k$}, it is simple to obtain that
\begin{small}
\begin{equation}
{\raise0.7ex\hbox{${{\partial ^2}{{\cal M}_k}}$} \!\mathord{\left/
{\vphantom {{{\partial ^2}{{\cal M}_k}} {\partial B_k^2}}}\right.\kern-\nulldelimiterspace}
\!\lower0.7ex\hbox{${\partial B_k^2}$}} = 0,
\end{equation}
\end{small}
which means that {\small $\mathcal{M}_k$} is linear to {\small $B_k$}.

For uplink transmit power {\small $P_k^{\left(U\right) }$}, let {\small $y \triangleq P_k^{\left(U\right) }$}. Using Proposition~\ref{berlemma} and substituting \eqref{id} into \eqref{faeg3}, we have
\begin{align}
E_k^{\left( U \right)}&\left( {\theta {y_1} + \left( {1 - \theta } \right){y_2}} \right) = \frac{{{\Gamma ^{ - 1}}\left( {{M_U}{N_Q}} \right)}}{{2\Gamma\!\left( {{\tau _2}} \right)\Gamma\!\left( {{M_C}{M_U}} \right)}}\frac{1}{{2\pi i}}
\notag\\&\times
\int_{\cal L} {\frac{{\Gamma\!\left( {{M_U}{N_Q} + t} \right)\Gamma\!\left( t \right)\Gamma\!\left( {t + {\tau _2}} \right)}}{{{\Gamma ^{ - 1}}\left( {{M_C}{M_U} - t} \right)\Gamma\!\left( {1 + t} \right)}}}
\notag\\&\times
{\left( {\frac{{\Lambda _k^{\left( U \right)}\left( {\theta {y_1} + \left( {1 - \theta } \right){y_2}} \right)}}{{{\tau _1}}}} \right)^t}{\rm{d}}t.
\end{align}
Using \eqref{delkuuse} in Lemma~\ref{delklemma}, we have
\begin{equation}
E_k^{\left( U \right)}\left( {\theta {y_1} \!+\! \left( {1\! -\! \theta } \right){y_2}} \right) \!\le\! \theta E_k^{\left( U \right)}\left( {{y_1}} \right)\!{{ + }}\!\left( {1 \!- \!\theta } \right)E_k^{\left( U \right)}\left( {{y_2}} \right)\!.
\end{equation}
Thus, {\small $E_k^{\left( U \right)}$} is convex to {\small $P_k^{\left(U\right) }$}. Because {\small $\mathcal{M}_k$} is linear to {\small $1-E_k^{\left( U \right)}$}, we derive that {\small $\mathcal{M}_k$} is concave in {\small $P_k^{\left(U\right) }$} to complete the proof.
\end{appendices}
\bibliographystyle{IEEEtran}
\bibliography{IEEEabrv,Ref}

\end{document}